\documentclass[journal]{IEEEtran}
\usepackage{cite}
\usepackage[cmex10]{amsmath}
\usepackage{amssymb}
\usepackage{amsfonts, amsthm}
\usepackage{array}
\usepackage[english]{babel}
\usepackage{graphicx}
\usepackage{float}
\usepackage{abbrevs}
\usepackage{cancel}
\usepackage{color}
\usepackage{tikz, subfigure}
\usepackage{soul}
\usepackage{multicol, blindtext}
\usepackage{cancel}
\usepackage{steinmetz}
\usepackage{epstopdf}
\usepackage{algorithm2e}
\usepackage{hyperref}
\newtheorem{theorem}{Theorem}
\newcommand{\partition}{\mathcal{P}}
\newcommand{\cell}{\mathbf{C}}
\newcommand{\sz}{\mathbf{z}}
\newcommand{\Prob}{\text{Pr}}

\begin{document}
	\title{{Poisson Multi-Bernoulli Mapping\\Using Gibbs Sampling}}
	\author{
		Maryam Fatemi, Karl Granstr\"{o}m, Lennart Svensson, Francisco J.\ R.\ Ruiz, and Lars Hammarstrand 
		\thanks{Maryam Fatemi was with the Department of Signals and Systems, Chalmers University of Technology, SE-412 96 Gothenburg, Sweden. She is now with Autoliv Sverige AB (email: maryam.fatemi@autoliv.com).
			Karl Granstr\"{o}m, Lennart Svensson and Lars Hammarstrand are with the Department of Signals and Systems, Chalmers University of Technology, SE-412 96 Gothenburg, Sweden (emails: \{karl.granstrom, lennart.svensson, lars.hammarstrand\}@chalmers.se).
		Francisco J.\ R.\ Ruiz is with the Department of Engineering, University of Cambridge, United Kingdom, and with the Computer Science Department, Columbia University, NY, USA (email: f.ruiz@columbia.edu).
		}}

	\maketitle
	\begin{abstract}
		This paper addresses the mapping problem. Using a conjugate prior form, we derive the exact theoretical batch multi-object posterior density of the map given a set of measurements. The landmarks in the map are modeled as extended objects, and the measurements are described as a Poisson process, conditioned on the map. We use a Poisson process prior on the map and prove that the posterior distribution is a hybrid Poisson, multi-Bernoulli mixture distribution. 
		 We devise a Gibbs sampling algorithm to sample from the batch multi-object posterior. The proposed method can handle uncertainties in the data associations and the cardinality of the set of landmarks, and is parallelizable, making it suitable for large-scale problems. The performance of the proposed method is evaluated on synthetic data and is shown to outperform a state-of-the-art method.
	\end{abstract}
	\section{Introduction}
	During recent years, self-driving cars have been the subject of extensive research. Many different functionalities expected from a self-driving vehicle are facilitated by having accurate localization and mapping capabilities. These capabilities require information of the environment which is gained by on-board sensors such as radars, cameras and internal sensors. The problem of mapping an unknown environment and estimating the unknown trajectory of the vehicle concurrently is referred to as simultaneous localization and mapping (SLAM) \cite{durrant2006}\cite{bailey2006}. SLAM has attracted a lot of attention, especially in the robotics community. Examples of SLAM algorithms are EKF-SLAM \cite{smith86} \cite{dissanayake2001}, FastSLAM \cite{montemerlo2002}, LMB-SLAM \cite{deusch2015} and Graph-SLAM \cite{grisetti2010}. The map produced by these methods is a statistical map of the environment as viewed by a particular sensor. To achieve higher positioning accuracy, it is common to separate the localization and mapping problems. In this case, the statistical mapping problem is solved offline using data from a vehicle equipped with a reference positioning system \cite{ziegler2014} \cite{lundgren2014} \cite{lundgren2015}. In such a system, if the position uncertainties are negligible, the focus can be on the statistical mapping.
	
	
	{Mapping has been addressed in, e.g.,  \cite{lundgren2015}, \cite{lundquist2011} and \cite{adams2014}.} Two aspects that make the problem challenging are the unknown data association between the measurements and the landmarks, and the unknown number of landmarks. In \cite{lundquist2011} and \cite{adams2014}, these two aspects are addressed through representing the map and the measurements as random finite sets (RFS), thus incorporating the uncertainties in the data association and the number of landmarks into the model. They use the probability hypothesis density (PHD) filter \cite{mahler2003multitarget} to recursively approximate the posterior density of the map. {In addition, \cite{lundquist2011} and \cite{adams2014} model the sensor detections as point objects, i.e., each object can generate at most one measurement at each time step. For some automotive sensors, the distance to landmarks is often such that a landmark is covered by more than one resolution cell of the sensor.} It is therefore more reasonable to use an extended object model, where each object can generate more than one measurement at each time step.
	
	A commonly used measurement model for extended objects is the inhomogeneous Poisson point process (PPP) proposed in \cite{gilholm2005}. In this model, a detected object generates a Poisson distributed number of measurements, and these measurements are spatially distributed around the object. This extended object model has been used in \cite{lundgren2015}, \cite{mahler2009phd}, \cite{orguner2011}, \cite{granstrom2012}. {In \cite{lundgren2015}, the authors present a vector-based extended object mapping, where a variational Bayesian expectation maximization (VBEM) algorithm \cite{beal2003} is used to approximate the posterior.} In the algorithm presented in \cite{lundgren2015}, the data associations and the map are jointly estimated; however, the uncertainty of the number of landmarks is not taken into account. 
		
In the mapping problem, where the landmarks are assumed to be extended objects, we need to cluster together the measurements that were generated by the same landmark in order to infer the properties of that landmark. However, the number of clusters is unknown because it corresponds to the number of landmarks. This resembles the problem addressed in the Bayesian nonparametric (BNP) literature \cite{hjort2010} \cite{gershman2012}. The BNP approach consists in fitting a model that can adapt its complexity to the observed data \cite{gershman2012}. In particular, the Dirichlet process (DP)\cite{ferguson1973} defines a prior distribution over an infinite-dimensional parameter space, ensuring that only a finite number of parameters are needed to generate any finite dataset. The posterior distribution reveals the number of parameters (clusters) that best fit the data. Both Markov Chain Monte Carlo (MCMC) methods \cite{robert2013} \cite{Neal2000} and variational methods \cite{jordan1999} have been proposed to perform inference in this context, where the former is the most widely used method of inference \cite{gershman2012}.  	
	
In this paper, we develop an MCMC method to estimate sensor maps that contain extended objects. {The proposed method can be applied to any sensor that is subject to false alarms, missed detections, and data association uncertainty.} The prior of the map is modeled by a Poisson process which can incorporate the uncertainty regarding the number of landmarks in a systematic manner, which gives important advantages compared to, e.g., the method of \cite{lundgren2015}, where the prior of the map does not incorporate uncertainties in the number of landmarks. Therefore, our method is capable of handling the uncertainties in both data associations and unknown number of landmarks. We derive the exact batch multi-object posterior density of the map using the conjugate prior form described in \cite{granstrom2016gamma}.\footnote{{An extended version of this paper was presented in \cite{granstromArxiv2016}, where the proof of conjugacy is included.}} To estimate the posterior, we derive a {collapsed Gibbs sampling \cite{robert2013} \cite{Neal2000}} algorithm which is analogous to widely used Gibbs sampling methods in BNP. We show that our method has the advantage of being parallelizable, making it suitable for large scale problems.

The paper is organized as follows. The problem formulation is presented in Section \ref{sec:problemFormulation}. Because in our derivations we make extensive use of the results given in \cite{granstrom2016gamma}, Section \ref{sec:sequentialPosterior} presents a background on this paper. The batch posterior density is derived in Section \ref{sec:posteriorDensities} followed by the proposed sampling method in Section \ref{sec:sampling}, where the similarities and differences between an alternative BNP approach and the proposed method are discussed. The inference algorithm is evaluated on a simulated scenario in Section \ref{sec:simulations} and conclusions are made in Section \ref{sec:conclusions}.

\section{Problem Formulation}
\label{sec:problemFormulation}
We are interested in mapping an area of interest (AOI). A vehicle equipped with a {sensor} navigates the AOI and collects data at discrete time steps, and our task is to construct a statistical model of those measurements as a function of the sensor pose. At each time instance the sensor only measures within its limited field of view (FOV) and the union of all these FOVs outlines the complete observed area (OA). In some scenarios the vehicle does not visit all parts of the AOI, in which case the OA is a subset of the AOI. Because of this, it is necessary to reason about both the parts of the AOI that have been observed, and the parts of the AOI that have not been observed.

The AOI is assumed to contain objects, here called landmarks. The map is the set of landmarks described by an RFS $\mathbf{\Theta}$, for which the cardinality and the properties (distribution) of its members are unknown. Each landmark is modelled as an elliptical extended object described by the landmark state $\mathbf{\theta}$ containing its 2D position $\mathbf{\mu}$, extent modelled as a random matrix $\mathbf{\Sigma}$, and expected number of measurements $\omega$. 

Mapping of the AOI is to estimate the batch posterior density of the set of landmarks given the set of collected measurements from $K$ time steps, i.e., $f(\mathbf{\Theta}|\mathbf{Z})$, where $\mathbf{Z} = \mathbf{Z}_1\bigcup \mathbf{Z}_2 \cdots \bigcup \mathbf{Z}_K$ is the union of $K$ sets of measurements. These measurements are modelled by the standard PPP extended object measurement model. Since the OA is a subset of the AOI, it is suitable to model the set of landmarks as the union of two disjoint subsets: a set of detected landmarks $\mathbf{\Theta}^d$ and a set of undetected landmarks $\mathbf{\Theta}^u$. The detected landmarks are within the OA, while the undetected ones are mostly outside the OA. However, because of detection uncertainty, there is a non-zero probability that there is one or more undetected landmarks inside the OA. The batch posterior density, denoted by $f(\mathbf{\Theta}|\mathbf{Z})$, is a Poisson MBM (PMBM) density. For the standard PPP extended object measurement model \cite{gilholm2005}, this is a conjugate prior \cite{granstrom2016gamma}.


In what follows, we describe the models and assumptions based on which we derive $f(\mathbf{\Theta}|\mathbf{Z})$ in Section \ref{sec:posteriorDensities}. The models and
derivations are based on finite set statistics \cite{mahler2007}. RFS based methods \cite{mahler2007} facilitate statistical inference in problems where the variables of interest and/or the measurements are modelled as finite sets. These methods are specially attractive because they can accommodate uncertainties in both the number of variables and their states. These two aspects match the statistical mapping problem very well since we are interested in estimating the number of landmarks as well as their properties. A brief overview of some key concepts of the random set theory required for following the presented models and the derivations, such as the distribution of a set-valued variable and its probability generating functional (p.g.fl), is given in Appendix \ref{sec:RFSbackground}. 
 
\subsection{Standard extended object measurement model}
\label{subsec:measurementModel}
In the standard extended object measurement model  \cite[pp. 431-432]{mahler2007} \cite{gilholm2005}, it is assumed that the measurements form an RFS that is the union of two independent RFSs: one for landmark generated measurements, and one for clutter. The clutter is modelled by a PPP with rate $\lambda_c$ and spatial density $c(\mathbf{z})$.
The p.g.fl of the clutter measurement model is
\begin{align}
G^c_k[g] &= \exp(\lambda_c\langle c; g \rangle -\lambda_c) \nonumber \\
& = \exp(\langle \kappa; g \rangle - \langle \kappa; 1 \rangle),
\label{eq:pgfClutter}
\end{align}  
where $\kappa(\mathbf{z})=\lambda_c c(\mathbf{z})$ is the clutter PPP intensity.

A landmark is detected at time $k$ with state dependent probability of detection $p_D^k(\mathbf{\theta})$. Measurements generated from landmarks are modelled by a PPP with rate $\gamma_k(\mathbf{\theta})$ and spatial distribution $\phi(\mathbf{z}_k|\mathbf{\theta})$. The conditional set likelihood of extended landmark measurements is expressed as \cite{granstrom2016gamma}
\begin{align}
\label{eq:measurementModel}
l_{\mathbf{z}_k}(\mathbf{\theta})&= p_D^k(\mathbf{\theta})p(\mathbf{Z}_k|\mathbf{\theta}) \nonumber \\
&= p_D^k(\mathbf{\theta})e^{-\gamma_k(\mathbf{\theta})} \prod_{\mathbf{z}_k\in\mathbf{Z}_k}\gamma_k(\mathbf{\theta})\phi(\mathbf{z}_k|\mathbf{\theta}).
\end{align}
The Poisson rate $\gamma_k(\mathbf{\theta})$ models the expected number of detections from a landmark with state $\mathbf{\theta}$.

The p.g.fl of the measurement model of a single extended landmark is
\begin{align}
G_k^l[g|\mathbf{\theta}] &= 1-p_D^k(\mathbf{\theta})+p_D^k(\mathbf{\theta})\exp(\gamma \langle \phi ; g \rangle -\gamma)
\end{align}
where $\langle \phi ; g \rangle=\int \phi(\mathbf{z}|\mathbf{\theta})g(\mathbf{z})d\mathbf{z}$ is a function of $\mathbf{\theta}$. It is assumed that the measurements originated from different landmarks are independent, therefore, the p.g.fl of the measurement model of multiple landmarks can be described by
\begin{align}
G_k^l[g|\mathbf{\Theta}]&= \left(1-p_D^k(\mathbf{\theta})+p_D^k(\mathbf{\theta})\exp(\gamma \langle \phi ; g \rangle -\gamma)\right)^{\mathbf{\Theta}}.
\label{eq:PGFLmeasurementMultiple}
\end{align}
Using (\ref{eq:pgflUnion}), (\ref{eq:pgfClutter}) and (\ref{eq:PGFLmeasurementMultiple}) the p.g.fl for the complete set of measurements can be expressed by
\begin{align}
G_k[g|\mathbf{\Theta}] &= G_k^c[g]G_k^l[g|\mathbf{\Theta}].
\end{align} 

\subsection{Modelling assumptions}
The prior RFS density of the map is described by a PPP. The map is assumed to be static, i.e., the appearance of landmarks on the map or their disappearance from the map is not accounted for in this work.

It is further assumed that the measurement batch contains measurements from $K$ time steps, and is denoted by the set $\mathbf{Z}$. A data association hypothesis over $\mathbf{Z}$ describes which measurements belong to which landmark, and partitions this set into non-empty subsets, called cells, whose members belong to the same landmark. The $i$th cell in partition $j$ is denoted by $\mathbf{C}^{j,i}$, further,
\begin{align}
\biguplus_{i} \{\mathbf{C}^{j,i}\} = \mathbf{Z} , \forall j.
\end{align}
In each cell we group the measurements by their time tags. Measurements with time tag $k$ form a set denoted by $\mathbf{V}^{j,i}_k$, therefore, 
\begin{align}
\biguplus_k  \{\mathbf{V}^{j,i}_k\}= \mathbf{C}^{j,i}, \forall j,i.
\end{align}
Note that $\mathbf{V}^{j,i}_k$ could possibly be empty for some $k$, as each landmark is within the FOV for a limited number of time steps.

In the clutter model described by (\ref{eq:pgfClutter}), we assume that clutter measurements are uniformly distributed over the field of view of the sensor, i.e.,
\begin{align}
c(\mathbf{z}) = {V}^{-1},
\end{align}
where $V$ is the observation volume of the sensor. In (\ref{eq:measurementModel}), we model
 \begin{align}
  \gamma_k(\mathbf{\theta}) =\omega f(\mathbf{\mu},\mathbf{x}_k)
 \end{align}
 where $\omega$ is the expected number of detections generated by a landmark, {$\mathbf{\mu}$ is the 2D position of the landmark,} $\mathbf{x}_k$ is the pose of the sensor at time $k$, and $f(\mathbf{\mu},\mathbf{x}_k)$ is the FOV function. This function is equal to one if $\mathbf{\mu}$ is in the FOV at time $k$, and zero otherwise. The connection between the sensor's field of view and the extent of a landmark is not taken into consideration for simplicity. This implies that a landmark is assumed to be outside the FOV if its centre is outside the FOV, even if a part of its extent is inside.  The spatial distribution is modelled as 
 \begin{align}
 \phi(\mathbf{z}_k|\mathbf{\theta}) = \mathcal{N}(\mathbf{z}_k;\mathbf{\mu},\mathbf{\Sigma}),
 \label{eq:singleMeasurement}
 \end{align}
 where $\mathbf{\Sigma}$ is a random matrix that describes the landmark extent. In addition, this model assumes that landmark measurements are generated by the centroid of the landmark, and since in reality the measurements could have been generated from any scattering point on the landmark, the measurement error corresponds to the extent of the landmark \cite{koch2008}. In this model it is assumed that the sensor measurement noise is negligible compared to a landmark's extent. {That is, if $\mathbf{\Sigma}_0$ is the true extent and $\mathbf{R}$ the measurement noise covariance, it is assumed that $\mathbf{R}$ is small compared to $\mathbf{\Sigma}_0$. More specifically, the measurement covariance in (\ref{eq:singleMeasurement}) is the added effect of $\mathbf{\Sigma}_0$ and $\mathbf{R}$, and is approximated as $\mathbf{\Sigma}=\mathbf{\Sigma}_0 + \mathbf{R} \approx \mathbf{\Sigma}_0$}. 
 

\section{Background}
\label{sec:sequentialPosterior}
In this section we present a summary of the sequential posterior RFS density for multiple extended objects. The reader is referred to \cite{granstrom2016gamma} and \cite{granstromArxiv2016} for a comprehensive description.

In \cite{granstromArxiv2016}, the authors prove that a Poisson multi-Bernoulli mixture (PMBM) RFS density is a conjugate prior for the Poisson extended object measurement model (presented in {Section} \ref{subsec:measurementModel}). In addition, it is shown that updating a Poisson prior with this measurement model results in a PMBM posterior. 
Therefore, for a batch problem, if the prior is a Poisson process, the posterior resulting from the batch update would be PMBM.
	
	In this section the results of \cite{granstrom2016gamma} are summarized. Note that only the results for the update step are presented here, this is because in this paper we are dealing with a static state (map) and there is no need to perform a prediction step. That is, it can be said that the predicted density is equal to the posterior density of the previous time step. 
	
	Similar to Section \ref{sec:problemFormulation}, the set of landmarks $\mathbf{\Theta}$ is divided into two disjoint subsets of detected and undetected landmarks denoted by $\mathbf{\Theta}^d$ and $\mathbf{\Theta}^u$, respectively. It is assumed that the p.g.fl of the prior density at time $k+1$ is
	\begin{align}
	G_k[h]&= G^u_k[h]G^d_k[h],
	\end{align}
	where 
	\begin{align}
	G^u_k[h]& = \exp{\left(\lambda^u_k \langle f^u_k;h\rangle -\lambda^u_k\right)}  \\
	G^d_k[h] &= \sum_{j=1} ^{N_k} W^j_k  \prod _{i=1}^{I_k^j} (1-r_k^{j,i}+r_k^{j,i} \langle  f_k^{j,i}; h\rangle), \label{eq:pgflD}
	\end{align}
	the Poisson rate of the undetected landmarks at time $k$ is denoted by $\lambda^u_k$, and $f^u_k$ is the spatial density of undetected landmarks at time $k$. The p.g.fl of detected landmarks is an MBM where the weight of the $j$th multi-Bernoulli component is denoted by $W^j_k$. This component corresponds to a partitioning of the set $\mathbf{Z}_k$ that consists of $I_k^j$ cells each denoted by $\mathbf{C}^{j,i}$. A cell refers to a subset of $\mathbf{Z}_k$ that cannot be empty. Accordingly, $N_k$ is the number of ways that the set $\mathbf{Z}_k$ can be partitioned into different cells, and it is equal to the Bell number of order $|\mathbf{Z}_k|$ \cite{Rota1964}. In each partition, measurements assigned to the same cell are assumed to have originated from the same source, i.e., the same landmark or clutter. The probability of existence of a landmark corresponding to cell $\mathbf{C}^{j,i}$ is $r_k^{j,i}$ and its spatial density is denoted by $f_k^{j,i}$. 
	
	The update step has two parts, updating the undetected landmarks, and updating the previously detected ones. Upon receiving the measurements at time $k+1$, the set $\mathbf{Z}_{k+1}$ is split into two subsets, namely, measurements not generated by previously detected landmarks $\mathbf{Y}_{k+1}$, and measurements generated by previously detected landmarks $\mathbf{Z}_{k+1}\backslash \mathbf{Y}_{k+1}$. The newly detected landmarks will form a MB RFS, {conditioned on a partitioning $\partition$} of the set $\mathbf{Y}_{k+1}$, 
	{
	\begin{align}
	G^{\partition}_{k+1}[h]&= \prod_{\Upsilon \in \partition} (1-r_{\Upsilon}+r_{\Upsilon}\langle f_{\Upsilon}; h \rangle),
	\end{align} }
	where 
	\begin{align}
	r_{\Upsilon}&= \left \{ \begin{array}{ccc}
	1 & \text{if} & |\Upsilon|>1  \\
	\frac{\mathcal{L}_{\Upsilon}}{\kappa^{\Upsilon}+\mathcal{L}_{\Upsilon}} & \text{if} & |\Upsilon|=1
	\end{array}\right. \label{eq:PeNew}\\
	f_{\Upsilon}(\mathbf{\theta})&= \frac{l_{\Upsilon}(\mathbf{\theta})\lambda^u_k f_{k}^u(\mathbf{\theta})}{\lambda^u_k \langle f_{k} ; l_{\Upsilon} \rangle} \label{eq:fNew} \\
	\mathcal{L}_{\Upsilon} &=\lambda^u_k \langle f_{k} ; l_{\Upsilon} \rangle.
	\end{align}
	Here, a cell of the partition {$\partition$} is denoted by $\Upsilon$, and $|\Upsilon|$ denotes the cardinality of the set $\Upsilon$. {In addition, $l_{\Upsilon}$ is the conditional set likelihood described in \eqref{eq:measurementModel}.}  
	The rate and spatial density of the PPP process for undetected landmarks are expressed by
	\begin{align}
	\lambda^u_{k+1}&= \lambda^u_{k} \langle f_{k}^u ; q_D^{k+1} \rangle \label{eq:rsU} \\
	f_{k+1}^u(\mathbf{\theta}) & = \frac{q_D^{k+1} (\mathbf{\theta})f_{k}^u(\mathbf{\theta})}{\langle f_{k}^u ; q_D^{k+1} \rangle} \label{eq:fsU} \\
	q_{D}^{k+1} (\mathbf{\theta}) & = 1- p_D^{k+1}(\mathbf{\theta})+p_D^{k+1}(\mathbf{\theta}) e^{-\gamma_{k+1}(\mathbf{\theta})},
	\end{align} 
	where $q_D^{k+1}$ is the effective probability of missed detection at time $k+1$, $p_D^{k+1}$ is the probability of detection and  $\gamma_{k+1}(\mathbf{\theta})$ is the intensity of the PPP process of the measurements.
	
	The set $\mathbf{Z}_{k+1} \backslash \mathbf{Y}_{k+1}$ contains measurements originated from the previously detected landmarks, and is partitioned into disjoint, possibly empty, subsets $\mathbf{V}_{k+1}^{j,i}$. The multi-Bernoulli component corresponding to this set is described by
	\begin{align}
	G_{k+1|k+1}^{\{\mathbf{V}_{k+1}^{j,i}\}_i}[h]&= \prod _{i} (1-r^{j,i}_{k+1}+r^{j,i}_{k+1} \langle  f^{j,i}_{k+1}; h\rangle)
	\end{align}
	where $\biguplus_i \{\mathbf{V}_{k+1}^{j,i}\}=\mathbf{Z}_{k+1} \backslash \mathbf{Y}_{k+1}, \forall j$. In addition, the probabilities of existence, the spatial distributions and the predicted likelihoods depend on the cardinality of the disjoint subsets. That is, if $\mathbf{V}_{k+1}^{j,i}\neq \emptyset$,
	\begin{align}
	r^{j,i}_{k+1}&=1 \label{eq:PeAgain} \\
	f^{j,i}_{k+1}(\mathbf{\theta})&=  \frac{l^{j,i}_{k+1}(\mathbf{\theta}) f_{k}^{j,i}(\mathbf{\theta})}{\langle f_{k}^{j,i} ;l^{j,i}_{k+1} \rangle} \label{eq:fAgain} \\
	\mathcal{L}_{k+1}^{\mathbf{V}_{k+1}^{j,i}} & = r_{k}^{j,i} \langle f_{k}^{j,i}; l_{k+1}^{j,i} \rangle . \label{eq:likelihoodAgain}
	\end{align}
	If $\mathbf{V}_{k+1}^{j,i}=\emptyset$,
	\begin{align}
	r^{j,i}_{k+1}&=\frac{r^{j,i}_{k}\langle f_{k}^{j,i} ; q_D^{k+1} \rangle}{1-r^{j,i}_{k}+r^{j,i}_{k}\langle f_{k}^{j,i} ; q_D^{k+1} \rangle} \label{eq:PeEmpty} \\
	f^{j,i}_{k+1}(\mathbf{\theta}) & = \frac{q_D^{k+1} (\mathbf{\theta})f_{k}^{j,i}(\mathbf{\theta})}{\langle f_{k}^{j,i} ; q_D^{k+1} \rangle} \label{eq:fEmpty} \\
	\mathcal{L}_{k+1}^{\mathbf{V}_{k+1}^{j,i}} & = 1-r_{k}^{j,i}+ r_{k}^{j,i} \langle f_{k}^{j,i}; q_D^{k+1} \rangle. \label{eq:likelihoodEmpty}
	\end{align}
	{In \eqref{eq:PeAgain} the existence probability of object $i$ in global hypothesis $j$ is equal to one. However, note that global hypothesis $j$ is not certain, but has a probability $W_k^j \in [0,1]$. Consequently, \eqref{eq:PeAgain} merely implies that object $i$ exists with probability 1 conditioned on hypothesis $j$.}
	
	The weight of each multi-Bernoulli component is computed by the predicted partition likelihood
	\begin{align}
	\mathcal{L}_P &= \prod_{\Upsilon \in P  |\Upsilon|>1} \mathcal{L}_{\Upsilon} \times \prod_{\Upsilon \in P  |\Upsilon|=1} (\kappa^{\Upsilon}+\mathcal{L}_{\Upsilon}),
	\end{align}
	and the predicted MB likelihood
	\begin{align}
	\mathcal{L}_{k+1}^{\{\mathbf{V}^{j,i}_{k+1}\}_i} & = \prod_{i}\mathcal{L}_{k+1}^{\mathbf{V}^{j,i}_{k+1}}. 
	\end{align}
	
	\section{Batch Multi-Object Posterior Density}
	\label{sec:posteriorDensities}
In this section, we present the exact theoretical multi-object batch posterior density. This distribution is calculated by performing a batch update using all the collected measurements. Naturally, we expect the batch posterior to be the same as the sequential posterior after having updated the latter with all collected measurements.

\subsection{The batch update}
The resulting batch multi-object posterior density has a PMBM form where the PPP part describes the posterior of the undetected landmarks and the MBM part describes the posterior of the detected landmarks and the clutter. The multi-object batch density and the update for parameters of the PPP and the MBM are described in Theorem \ref{th:batchPosterior}, the proof of which is presented in Appendix \ref{sec:appendix}.
\begin{theorem}
	\label{th:batchPosterior}
	The multi-object batch posterior density of the map $\mathbf{\Theta}$ given the set of measurements $\mathbf{Z}$ has a PMBM form and is expressed as
	\begin{align}
	f(\mathbf{\Theta}) &= \sum_{{\mathbf{\Theta}}^d \subset {\mathbf{\Theta}}} \left(\sum_{j=1}^{N_K} W^j_K \sum_{\alpha^j \in P^{|\mathbf{\Theta}^d|}_{I_K^j}} \prod_{i=1}^{I_K^j} f^{j,i}_K(\mathbf{\Theta}_{\alpha^j(i)}) \right) \times \nonumber \\
	& \left( e^{-\lambda_K^u} \prod_{\mathbf{\theta}\in (\mathbf{\Theta} \backslash \mathbf{\Theta}^{d})} \lambda_K^u f_K^u(\mathbf{\theta}) \right),
	\label{eq:posterior}
	\end{align}
	 where the first factor describes an MBM, the second factor is a PPP and the entire set $\mathbf{\Theta}$ is the union of these disjoint sets. In the MBM, the index $j$ describes a partitioning of an entire set of the measurements and represents an MB in the mixture, whereas, the index i corresponds to different Bernoulli components in an MB. The mapping $P^{|\mathbf{\Theta}^d|}_{I_K^j}$, defined in (\ref{eq:mappingMBM}),  accounts for all different ways of assigning each Bernoulli of each multi-Bernoulli component to a landmark for different number of landmarks, and $\alpha^j$ is a function that maps each Bernoulli component of the $j$-th multi-Bernoulli into either a landmark $l \in \{1,2,\ldots,{|\mathbf{\Theta}^d|}\}$ or clutter. In addition, $\mathbf{\Theta}_{\alpha^j(i)}$ is described by (\ref{eq:componenrMBM}) {and the index $K$ denotes the final time step}.
	 	
	 The parameters of the $(j,i)$-th Bernoulli component of the MBM density corresponding to the set $\mathbf{Z}$, the cell $\mathbf{C}^{j,i}$ and subsets $\mathbf{V}_k^{j,i}$ are given by
	 \begin{align}
	 	f_K^{j,i}(\mathbf{\theta}) &= \frac{f_{0}^u(\mathbf{\theta}) 
	 		\prod_{k=1}^K l^{j,i}_{k}}{\langle f_{0}^u ; \prod_{k=1}^K l^{j,i}_{k} \rangle }
	 	\label{eq:fBernoulliD} \\
	 	r_K^{j,i} & = \left\{ \begin{array}{ccc}
	 	\frac{{L}_{K}^{j,i}}{\lambda_c c(\mathbf{z})+{L}_{K}^{j,i} } & \text{if} & |\mathbf{C}^{j,i}|=1 \\
	 	1 & \text{if} & |\mathbf{C}^{j,i}|>1
	 	\end{array}\right. , \label{eq:rBernouliD}
	 \end{align}
	 where
	 \begin{align}
	 l_k^{j,i} &= \left \{\begin{array}{cc}
	 p_D^{k}(\mathbf{\theta}) e^{-\gamma_k(\mathbf{\theta})}\prod_{\mathbf{z} \in V^{j,i}_{k} } {\gamma_{k}(\mathbf{\theta}) \phi(\mathbf{z}|\mathbf{\theta})} & \mathbf{V}_k^{j,i} \neq \emptyset \\
	 q_D^k(\mathbf{\theta}) & \mathbf{V}_k^{j,i}=\emptyset 
	 \end{array} \right.  \label{eq:likelihoodbothEmptyandNot}\\
	 	{L}_{K}^{j,i} & = \lambda_{0}^u \langle f_{0}^u ;	\prod_{k=1}^K l^{j,i}_{k} \rangle \label{eq:L}
	 \end{align}
	 and $q_{D}^k (\mathbf{\theta})  = 1- p_D^{k}(\mathbf{\theta})+p_D^{k}(\mathbf{\theta}) e^{-\gamma_k(\mathbf{\theta})}$, is the effective probability of missed detection at time $k$. 
	 	The weight of each multi-Bernoulli component, $W_K^j$ is expressed as
	 	\begin{align}
	 	W_K^j & \propto \prod_{i} \mathcal{L}_K^{j,i}
	 	\label{eq:weights}
	 	\end{align}
	 	where $\mathcal{L}_K^{j,i}$ is the likelihood of the MB which corresponds to cell $i$ in partition $j$. This likelihood depends on $|\mathbf{C}^{j,i}|$ and is described by
	 	\begin{align}
	 	\mathcal{L}_K^{j,i}& = \left \{ \begin{array}{cc}
	 	\lambda_c c(\mathbf{z}) + {L}_{K}^{j,i} & |\mathbf{C}^{j,i}|=1 \\
	  {L}_{K}^{j,i} & |\mathbf{C}^{j,i}|>1
	 	\end{array}  \right..
	 	\end{align}
	 	
	 	 The PPP, which corresponds to the density of undetected landmarks, does not depend on the measurements.  
	 	 The Poisson rate and the spatial density of the PPP are expressed as
	 	 \begin{align}
	 	 \lambda_K^u & = \lambda_0^u \langle f^u_{0};\prod_{k = 1}^{K}q_D^k \rangle  \label{eq:rateU}\\
	 	 f^u_K(\mathbf{\theta}) & =  \frac{ f^u_0(\mathbf{\theta}) \prod_{k=1}^{K} q_D^k(\mathbf{\theta})}{\langle f^u_{0} ; \prod_{k=1}^{K}q_D^k\rangle} . \label{eq:fU}
	 	 \end{align}
	 	
\end{theorem}


In the MBM, each partition of the measurements indexed by $j$ represents an MB. Each cell in the $j$th partition (MB) is denoted by $\mathbf{C}^{j,i}$, corresponds to a Bernoulli component, and is assumed to contain measurements belonging to the same source (landmark). The probability of existence, expressed in (\ref{eq:rBernouliD}), is one for the Bernoulli components that have been assigned more than one measurement, i.e., the Bernoulli components which correspond to $|\mathbf{C}^{j,i}|>1$. This is because, the clutter process generates independent point measurements which implies that two clutter measurements cannot be assigned to the same cell. Accordingly, the probability of existence for Bernoulli components associated with a single measurement is less than one, since it is not certain whether this measurement is generated by clutter or a landmark. 
According to (\ref{eq:fBernoulliD}), the spatial density of each Bernoulli component of the MBM originates from the spatial density of an undetected landmark, and is calculated by updating the spatial density of the undetected landmark with all the measurements assigned to the corresponding Bernoulli component. Additionally, the weight of each MBM, described in (\ref{eq:weights}), is proportional to a product of the likelihood of each cell. 

In (\ref{eq:rateU}), the term $ \langle f^u_{0};\prod_{k = 1}^{K}q_D^k \rangle$ is the probability that a landmark with density $f_0^u(\mathbf{\theta})$ is not detected from time 1 to $K$. The effective probability of missed detection at time $k$ is $q_D^k$, which according to (\ref{eq:likelihoodbothEmptyandNot}) could also be regarded as the likelihood of an empty measurement set at time $k$. This probability is 1 outside the FOV and lower inside. The rate of the PPP describing undetected landmarks is decreased inside the union of the FOVs that have been surveyed by the sensor. This rate should be lowest over the areas which have been surveyed the most. Outside the union of the surveyed FOVs, $\lambda_K^u$ does not change.
The spatial density of the undetected landmarks given by (\ref{eq:fU}) expresses that this density does not change outside the union of the FOVs. Furthermore, the changes to the spatial density inside the FOVs indicate that the expected number of detections, parametrized in this density through $\omega$, should be decreased inside the FOVs.

\subsection{Approximations}
A major difficulty in computing the PMBM density in (\ref{eq:posterior}) is the second summation which has $N_K$ terms, i.e., the Bell number of order $|\mathbf{Z}|$. Since the sequence of Bell numbers grows very fast and set $\mathbf{Z}$ corresponds to the measurements collected over many time steps, $N_K$ is possibly a very large number. For example, for only 10 measurements, $N_K=115975$. 

Given the level of complexity, computing the posterior density exactly is intractable and there is a need to incorporate appropriate approximations into the computations.
{ Approximating the posterior distribution is the central algorithmic problem of Bayesian inference. Most Bayesian inference algorithms can be grouped into variational methods and MCMC methods. Variational inference approximates the posterior with a simpler family of distributions, whose parameters are chosen to minimize the Kullback-Leibler divergence between the approximation and the true posterior \cite{jordan1999} \cite{Wainwright2008}. In contrast, in MCMC methods,} a Markov chain is defined over the hidden variables such that the equilibrium distribution of this chain is the true posterior {\cite{Liu2004} \cite{robert2013} \cite{Owen2013}}. Drawing samples from the chain eventually results in obtaining samples from the posterior. The advantage of MCMC methods is that the sample distribution is generally guaranteed to converge to the posterior {provided that we draw enough samples. Variational inference does not have such property \cite{Wainwright2008}.}


{In Section~\ref{sec:sampling}, we present an inference algorithm based on Gibbs sampling.} Gibbs sampling is a simple form of MCMC sampling where the Markov chain is constructed by sampling from conditional distributions of each hidden variable given the rest of the variables and the measurements.


\section{Sampling the partitions}
\label{sec:sampling}
In this section, we present an MCMC based inference method. As the parameter space of the map can be quite large, sampling from those parameters could result in a slow and inefficient inference algorithm. However, using certain assumptions, the model structure enables us to marginalize all the continuous variables analytically, which has been demonstrated to substantially improve the mixing properties of the Gibbs sampler \cite{Neal2000}. Therefore, inspired by the same situation in Dirichlet process inference methods which deal with the situation where the number of classes is unknown \cite{Neal2000}, we devise an efficient algorithm to sample from the distribution of the partitions, i.e., the distribution of the data association hypotheses over the complete batch of measurements. 

Each sample is a partition of the measurements which corresponds to an MB in the MBM described in (\ref{eq:posterior}). That is, each sample corresponds to a distribution of the detected landmarks. To calculate an estimated map, we perform an averaging over the landmarks present in the substantial number of the final samples of the chain (avoiding the burn-in samples).
	
In this section we focus on inferring the posterior density of the detected landmarks, i.e., 
\begin{align}
f_{\mathbf{\theta}^d}(\mathbf{\Theta}^d) = \sum_{j=1}^{N_K} W^j_K \sum_{\alpha^j \in P^n_{I_K^j}} \prod_{i=1}^{I_K^j} f^{j,i}_K(\mathbf{\Theta}_{\alpha^j(i)}),
\label{eq:posteriorDetected}
\end{align}
{where the index $j$ describes the partitioning $\partition$ of the entire set of data. We can now give $j$ a probabilistic interpretation, and observe that the joint distribution over the partition and the detected landmarks, described by
\begin{align}
f(\mathbf{\Theta}^d, j) &= W^{j}_K \sum_{\alpha^{j} \in P^{|\mathbf{\Theta}|^d}_{I_K^{j}}} \prod_{i=1}^{I_K^{j}} f^{{j},i}_K(\mathbf{\Theta}_{\alpha^{j}(i)}),
\end{align} }
 matches the marginal distribution in (\ref{eq:posteriorDetected}). 
 
Accordingly, the probability mass function of the partitions can be obtained by marginalizing out the detected landmarks,
{\begin{align}
\text{Pr}\{j \} & = W^{j}_K.
\label{eq:partitionWeights}
\end{align}}This distribution describes the probabilities of partitions of the measurements, i.e., the probabilities of different data association hypotheses over the batch of the measurements, which {are} equal to the weights of each multi-Bernoulli component $W^{j}_K$. 
The ability to compute the weights, at least up to a proportionality constant, is key to a sampling method that samples from the partitions.
		  
In this section a closed form solution for the computation of the weights (up {to} a proportionality constant) is described and a Gibbs sampler which can sample from the partitions is proposed. 
\subsection{Computing the weights}
Computing the weights in (\ref{eq:partitionWeights}) is possible if the integral in (\ref{eq:L}) has a closed form solution. To find a closed form solution the following assumptions have been made: the probability of detection is constant inside the field of view, and the parameters of the map are a priori distributed according to a conjugate prior form expressed by
\begin{align}
p(\mathbf{\mu}) & = \mathcal{U}(\mathbf{\mu}), \nonumber \\
p(\mathbf{\Sigma}) &= \mathcal{IW}(\mathbf{\Sigma};\mathbf{S}_0,\nu_0) \nonumber \\
p(\omega) &= \mathcal{GAM}(\omega;\alpha_0,\beta_0),
\label{eq:prior}
\end{align} 
where $\mathcal{U}(\mathbf{\mu})$ is a uniform distribution over the AOI. If we denote this region by $\mathcal{A}$, the uniform distribution can be expressed as $\mathcal{U}(\mathbf{\mu}) = \frac{\mathcal{I}(\mathcal{A})}{V_{\mathcal{A}}}$, where $\mathcal{I}(\mathcal{A})$ is an indicator function over $\mathcal{A}$ and $V_{\mathcal{A}}$ is the volume of this region. In addition, $\mathcal{IW}(\cdot)$ denotes an inverse Wishart density and $\mathcal{GAM}(\cdot)$ denotes a gamma distribution. Using the above assumptions, the calculation of the {weights} is tractable.

The integral in (\ref{eq:L}) corresponds to having updated parameters of a landmark with the measurements in cell $\mathbf{C}^{j,i}$, that is with multiple empty and non-empty subsets $\mathbf{V}_k^{j,i}$. The result of an update with multiple non-empty $\mathbf{V}_k^{j,i}$s is a normal inverse Wishart gamma distribution whose parameters have been updated by the measurements in these subsets.

 Empty $\mathbf{V}_k^{j,i}$s correspond to missed detection events. Such events only affect the parameters of the resulting gamma distribution. Each missed detection creates two hypotheses regarding the distribution of the expected number of detections from a landmark. The two hypotheses describe two different ways by which our measurement model accounts for an empty set of detections. One of the hypotheses corresponds to events such as occlusion in the FOV, in which case we do not need to change our prior belief regarding the expected number of detections from a landmark. The other hypothesis states that the landmark in the FOV has generated zero measurements thus we should expect fewer detections from this landmark, and this information should be used to update our prior belief.
 
 Accordingly, the result of an update by multiple empty and non-empty sets of measurements is a posterior with unimodal normal-inverse Wishart part, and multi modal gamma part. The gamma has $2^{N_{\emptyset}}$ modes, where $N_{\emptyset}$ is the number of empty sets in $\mathbf{C}^{j,i}$ for which $ f(\mathbf{\mu}, \mathbf{x}_k)=1$. The multimodal gamma distribution should be reduced to a unimodal gamma distribution, to avoid computational complexity issues. 
In this paper we have used a simple merging technique which works well for high $p_D$ and low true landmark Poisson rate. In scenarios with low $p_D$ and/or high landmark Poisson rate a  more accurate gamma mixture reduction technique, such as the merging technique based on Kullback-Leibler divergence \cite{kullback1951} presented in \cite{granstrom2012estimation}, can be used.

 Using the above assumptions, a landmark updated by the measurements in the cell $\mathbf{C}^{j,i}$  will have a normal gamma inverse Wishart distribution expressed as,
 \begin{align}
 \mathcal{N}(\mathbf{\mu};\mathbf{\mu}_K^{j,i},(c_K^{j,i})^{-1}\mathbf{\Sigma})\mathcal{GAM}(\omega;\alpha_K^{j,i},\beta_K^{j,i})\mathcal{IW}(\mathbf{\Sigma};\mathbf{S}_K^{j,i},\nu_K^{j,i}), \nonumber
 \end{align}
whose parameters are given by
 \begin{align}
 \nu_{K}^{j,i} &= \nu_0 + |\mathbf{C}^{j,i}|-1\nonumber \\
 \mathbf{S}_{K}^{j,i} &=  \mathbf{S_0} +  \sum_{\mathbf{z} \in \mathbf{C}^{j,i}} (\mathbf{z}-\overline{\mathbf{z}}) (\mathbf{z}-\overline{\mathbf{z}})^T  \nonumber \\
 \mathbf{\mu}_{K}^{j,i} &= \overline{\mathbf{z}}  \nonumber \\
 c_{K}^{j,i} &=  |\mathbf{C}^{j,i}| \nonumber \\
 \alpha_{K}^{j,i} & = \alpha_0 + |\mathbf{C}^{j,i}| \nonumber \\
 \beta_K^{j,i} &= \beta_0 + N_1 + \sum_{n=0}^{N_{\emptyset}} {{N_{\emptyset}}\choose{n}}(p_D)^n(1-p_D)^{N_{\emptyset}-n}n \nonumber \\ &= \beta_0 + N_{1} + p_D N_{\emptyset}
 \label{eq:batchUpdate}
 \end{align}
 where $\overline{\mathbf{z}}$ is the mean of the measurements in cell $\mathbf{C}^{j,i}$, and $\beta_K^{j,i}$ is calculated by a weighted sum over all the hypotheses due to missed detections under the assumption of having a constant probability of detection {for all landmarks in the FOV}, i.e., $p_D^k = p_D$ for all $k$. In addition, $N_1$ is the cardinality of a set of time stamps $k$ for which $\mathbf{V}_k^{j,i}\neq \emptyset$. 
 Accordingly the integral in (\ref{eq:L}) which corresponds to the weight of the $j$th partition of the measurements is expressed as 
 \begin{align}
 W_K^j & \propto  \prod_i \frac{\prod_{k: \mathbf{V}_k^{j,i}\neq \emptyset} p_D^{k}}{\mathbf{V}_{\mathcal{A}}}\frac{\beta_0^{\alpha_0} \Gamma(\alpha_{K}^{j,i} )}{(\beta_{K}^{j,i} )^{\alpha_{K}^{j,i}}\Gamma(\alpha_0)} \times \nonumber \\ &\frac{|\mathbf{S}_0|^{\nu_0/2}\Gamma_2(\nu_{K}^{j,i}/2)}{\pi^{|\mathbf{C}^{j,i}|-1}(c_{K}^{j,i})^{0.5}\Gamma_2(\nu_0/2)|\mathbf{S}_{K}^{j,i}|^{\nu_{K}^{j,i}/2}}.
 \label{eq:intBatch}
 \end{align}
The results in (\ref{eq:batchUpdate}) and (\ref{eq:intBatch}) are derived using similar mathematical derivations as \cite{koch2008} and \cite{granstrom2012phd}.

 In this section the probability mass function of the partition function was derived. We showed that the probability of each partitioning of the measurements is equal to $W^{j}_K$. In addition, it was shown that with certain assumptions regarding the prior densities and the likelihood of the measurements, these weights can be calculated up to a proportionality constant. In the remaining of the section, we propose a Gibbs sampler to sample from the partition function.

		 \subsection{A Gibbs sampler}	
		 \label{subsec:GibbsSampler}	 
		 The goal is to sample from the partitions. Each partitioning of the measurements corresponds to a different multi-Bernoulli component in the MBM. To sample from the partitions we need to devise a method that facilitates moving from one partition to the next. { In this section we present one such method. Let $\partition^{(t)}$ be the partition at the $t$-th iteration of the Gibbs sampler. Assume that $\partition^{(t)} = \partition^{(j)}$, where $\partition^{(j)}$ is a valid partition with cells $\cell^{j,i}$. The cells are indexed by $i\in\left\{1,2,\ldots,n_{\partition^{(j)}} \right\}$, where $n_{\partition^{(j)}} = |\partition^{(j)}|$ is the number of cells in partition $\partition^{(j)}$. Note that a partition is invariant to the ordering of the cells, i.e., if we permute the cell indices we have the same partition.
		 
		 The $(t+1)$-th partition is obtained as follows. First a measurement $\sz$ is randomly selected from the measurement set. Assume without loss of generality that the selected measurement belongs to the $\ell$-th cell, $\sz\in\cell^{j,\ell}$. New partitions can be obtained by performing ``actions'' that involve the selected measurement $\sz$. Specifically, we consider the following actions:
		 \begin{itemize}
		 	\item Move $\sz$ from $\cell^{j,\ell}$ to the $m$-th cell $\cell^{j,m}$, where $m\in\left\{1,2,\ldots,n_{\partition^{(j)}} \right\}$. We denote the resulting partition as $\partition^{(j^{\prime})}_{m}$.
		 	\item Move $\sz$ from $\cell^{j,\ell}$ to a new cell. We denote the resulting partition as $\partition^{(j^{\prime})}_{0}$.
		 \end{itemize}
		 The probability of moving to a partition is
		 \begin{align}
		 \Prob \left\{ \left. \partition^{(t+1)} = \partition^{(j^{\prime})}_{m} \right| \partition^{(t)} = \partition^{(j)},\sz \in \cell^{j,\ell} \right\} = \frac{ W^{ \partition^{ ( j^{\prime} ) }_{ m } }_K }{ \sum_{m^{\prime} = 0}^{ n_{ \partition^{(j)} }} W^{ \partition^{ ( j^{\prime} ) }_{ m^{\prime} } }_K } \label{eq:GibbsTransitionProbability}
		 \end{align}
		 for $m=0,1,\ldots,n_{\partition^{(j)}}$, where $W^{ \partition^{ ( j^{\prime} ) }_{ m } }$ is the weight of the partition \eqref{eq:intBatch}. Note that $m=0$ means that we get the same partition, i.e., this action corresponds to $\partition^{(t+1)} = \partition^{(t)}$. Further, for the special case $|\cell^{j,\ell}| = 1$, moving $\sz$ to a new cell results in an equivalent partition, i.e., if $|\cell^{j,\ell}| = 1$ and $m=\ell$ then $\partition^{(t+1)} = \partition^{(t)}$. Because of this, to avoid double counting the probability of $\partition^{(t+1)} = \partition^{(t)}$, if $|\cell^{j,\ell}| = 1$ we set
		 \begin{align}
		 \Prob \left\{ \left. \partition^{(t+1)} = \partition^{(j^{\prime})}_{\ell} \right| \partition^{(t)} = \partition^{(j)}, \sz \in \cell^{j,\ell} \right\} = 0.
		 \end{align}
		 
		 The algorithm can be initialized with any valid partition. Two simple examples are starting with a partition with all the measurements in individual cells, or starting with a partition with all measurements in a single cell. In the simulation section, we use the former alternative.
		 
		 The sum in the denominator of \eqref{eq:GibbsTransitionProbability} might pose a challenge as it could include a large number of cells. There are two relieving aspects for the calculation of this sum. First, we expect many of the terms to be zero as they could correspond to infeasible data associations. For example, if the time tag of $\mathbf{z}$ is $k$ and the landmark formed by the measurements in $C^{j,m}$ is out of the field of view at time $k$, then it can be concluded that the corresponding $W^{\mathcal{P}^{(j^{\prime})}_{m}}_K=0$. Such cases can be removed from the possible partitions.  Second, since each move only changes two cells, we can reuse the parameters calculated for the other cells when calculating the weight of the newly formed partition, i.e.,
		 \begin{align}
		 W^{\mathcal{P}^{(j^{\prime})}_{m}}_K & \propto \prod_{\tau} \mathcal{L}^{ \mathcal{P}^{(j^{\prime})}_{m},{\tau}}_K \nonumber \\
		 & = \mathcal{L}^{\mathcal{P}^{(j^{\prime})}_{m},{\ell}}_K \mathcal{L}^{\mathcal{P}^{(j^{\prime})}_{m},{m}}_K \prod_{\tau : \tau \neq \ell,m} \mathcal{L}^{\mathcal{P}^{(j^{\prime})}_{m},{\tau}}_K \nonumber \\
		 & = \mathcal{L}^{\mathcal{P}^{(j^{\prime})}_{m},{\ell}}_K \mathcal{L}^{\mathcal{P}^{(j^{\prime})}_{m},{m}}_K \prod_{\tau : \tau \neq \ell,m} \mathcal{L}^{\mathcal{P}^{(j)},{\tau}}_K
		 \label{eq:}
		 \end{align}
		 where $\mathcal{L}^{ \mathcal{P}^{(j^{\prime})}_{m},{\tau}}_K$ denotes the likelihood of cell $\tau$ of partition $ \mathcal{P}^{(j^{\prime})}_{m}$. Note that when a measurement is moved out of a cell, the cell might become empty and we define, $\mathcal{L}_{K}^{\partition_{m}^{(j^{\prime}),{\emptyset}}}\triangleq1$. }
		 \subsection{Parallelization}
		 The proposed sampling method can be parallelized. This can be viewed as a result of the product in (\ref{eq:weights}), limited FOV of the sensor and the size of the landmarks. For example, measurements that have been detected in two non-overlapping field of views have zero probability of belonging to the same landmark. This implies that we can process two such groups of measurements in parallel without having to introduce any approximations. Moreover, measurements that are far away (relative to our current belief of the size of the landmarks) will have very low probability of belonging to the same landmark. While this probability could be non-zero in practice, we can use this information to obtain further parallelization by gating those measurements. This results in an approximated but accelerated sampling procedure.
		 
		 
		 \subsection{Comparison to BNP}
		 One commonly used prior model in BNPs is the Dirichlet process \cite{ferguson1973}, which is typically applied as a building block in the Dirichlet process mixture model (DPMM). This model assumes that there is an infinite number of components, of which a finite number have generated the finite dataset at hand. This type of modelling results in a posterior distribution that assigns higher probability mass to the number of components that best explains the observations. This avoids the need to pre-specify the number of components before observing the data \cite{hjort2010} \cite{gershman2012}.
		 
		 In this paper, the map has been assigned a Poisson process prior. Similarly to the DPMM, our model allows us to infer the number of components (landmarks) given the data. The difference is that the Poisson prior assumes that there exists a finite and unknown number of landmarks a priori. 
		 
		 The Poisson prior model of the map is motivated by its favorable conjugate properties to the standard measurement model of extended landmarks. In \cite{granstromArxiv2016} it is shown that the PMBM prior is conjugate to the PPP extended target measurement model. In a sequential setting, at the first time step, our Poisson prior is updated by this measurement model, resulting in a PMBM posterior. Consequently, the posterior will maintain its PMBM form in the subsequent time steps. Similarly, in a batch setting the Poisson prior and the PPP measurement model result in a PMBM posterior. Therefore, while the Poisson model for the map ensures a closed form posterior, it is not clear how a Dirichlet process prior would shape the posterior given the PPP measurement model.
		 
		 The Gibbs sampling method used in this paper is similar to the methods discussed in \cite{Neal2000} \cite{maceachern1994}, where the associations of measurements to components are updated sequentially one at a time, conditioned on the data and the rest of the associations. This results in an MCMC chain whose equilibrium distribution is the posterior density of the component associations. This conditional probability can be viewed as the probability of a particular partitioning of the measurements, since one can move from one partitioning of the measurements to another by changing a component (landmark) assignment of one measurement. From this perspective, the equilibrium distribution of the MCMC chain is the posterior density of the measurements' partitions. In the proposed method we have derived the probability distribution of partitions of the measurements formed by changing the landmark assignment of one measurement. 
		 
		 \section{Estimation of the map}
		 \label{sec:mapEstimation}
		 In this section we describe how the result of the sampling method is used for estimating the map of the environment.
		 
		We assume that the chain has run long enough for convergence, as a result the final samples are samples of the posterior density of the partitioning of the measurements. That is, the number of sampling iterations should be chosen such that the burn-in period of the sampler is avoided. Given a partition of the measurements, the association of the measurements to landmarks is known. Therefore, each sample of the posterior corresponds to a map of the environment. The estimated map is computed by averaging over a number of final samples of the chain. The averaging is performed over the positions, extents and the weights of the landmarks of each sample (map). 
		
		In each partition of the measurements, every cell that contains more than one measurement is considered to have been generated by one landmark. This landmark has the prior distribution described in (\ref{eq:prior}) and is updated by the measurements assigned to it according to (\ref{eq:batchUpdate}). Among the cells which contain a single measurement, some are generated by landmarks. To decide whether a single measurement is generated by a landmark, a threshold is set on the probability of existence of the cells with cardinality one. Cells whose probability of existence are above the threshold are considered to have been generated by a landmark. Consequently, the parameters for these landmarks are updated by the single measurement assigned to them. According to (\ref{eq:batchUpdate}), the parameters of the extent of such landmarks are not updated, since for a uniform position prior a single measurement only contains information about the position and not the extent of the landmark. 
		
		To perform the averaging over the samples, we need to solve an association problem between the cells of the samples. That is, we need to establish which cells across the samples belong to the same landmark. This is done by specifying an area around the position of each cell. The cells whose distance is less than the specified area are assumed to belong to the same landmark. To estimate the properties of a landmark, we average over the properties of the cells that belong to that landmark. We do not use spurious cells in this calculation. Spurious cells are those that exist in a very small percentage of the chosen samples. The resulting estimation method is presented in Algorithm \ref{alg:mapping}. 
		
			\begin{algorithm} 
				\caption{Pseudo-code of the mapping algorithm.}\label{alg:mapping}
				
				\emph{\textbf{Gibbs Sampling}} \\
				Choose an initial partitioning $\mathcal{P}_I$ of the measurements, See Section \ref{sec:mapEstimation}.\\
				\For{n=1: \#iterations}{
					Randomly choose a measurement $\mathbf{z}$ from a cell. \\
					Evaluate the weight of possible partitions formed by moving $\mathbf{z}$ to a different cell according to Eq. (\ref{eq:intBatch}). \\
					Choose a partition $\mathcal{P}$ according to Eq. (\ref{eq:GibbsTransitionProbability}). \\
				}
				\emph{\textbf{Map Estimation}} \\
				Choose a number of final samples of the chain, see Section \ref{sec:mapEstimation}.\\
				\For{j=1: \#Samples}
				{sample $j$ corresponds to partition $\mathcal{P}$. \\
					\For{i=1:\#Cells in $\mathcal{P}$ }
					{ \If{$r^{\mathcal{P},i}>$threshold}
						{update the parameters of the corresponding landmark according to Eq. (\ref{eq:batchUpdate}).}
					}
				}	
				Average over the landmarks of the samples, see Section \ref{sec:mapEstimation}.
				\vspace{1mm}
				\end{algorithm}
		
		The initial partitioning of the measurements can be chosen arbitrarily so long as the chosen partition is feasible. An infeasible partition corresponds to a partition with at least one cell containing measurements from spatially far apart areas of the map. Because the measurements in the same cell are hypothesized to belong to the same landmark, such a partition is infeasible and has zero weight. A simple choice of an initial partitioning of the measurements is a partition with single-member cells where each measurement belongs to its own cell.
		
		The map generated by each sample of the MCMC chain can be viewed as the intensity function of the PPP that has generated the measurements. This per sample map is later used for performance evaluation in Section \ref{sec:simulations}. The intensity function is an unnormalized Gaussian mixture. The mixture corresponding to the estimated map is given by
		\begin{align}
		\zeta(\hat{\mathbf{\theta}})&=\sum_{l=1}^{N_E} \hat{\omega_l} \mathcal{N}(\mathbf{\mu}_l;\hat{\mathbf{\mu}}_l,\hat{\mathbf{\Sigma}}_l),
		\end{align}
		where $\hat{\mathbf{\theta}}$ is formed by choosing an arbitrary order of the estimated landmarks, and $N_E$ is the estimated number of landmarks. Additionally, $\hat{\omega_l}$, $\hat{\mathbf{\mu}}_l$ and $\hat{\mathbf{\Sigma}}_l$ are the mean of the updated gamma, normal and inverse Wishart distribution of landmark $l$, respectively. 	 

		 \section{Simulations and Results}
		 \label{sec:simulations}
\begin{figure}
		 	\centering
		 	\includegraphics[width=0.5\textwidth]{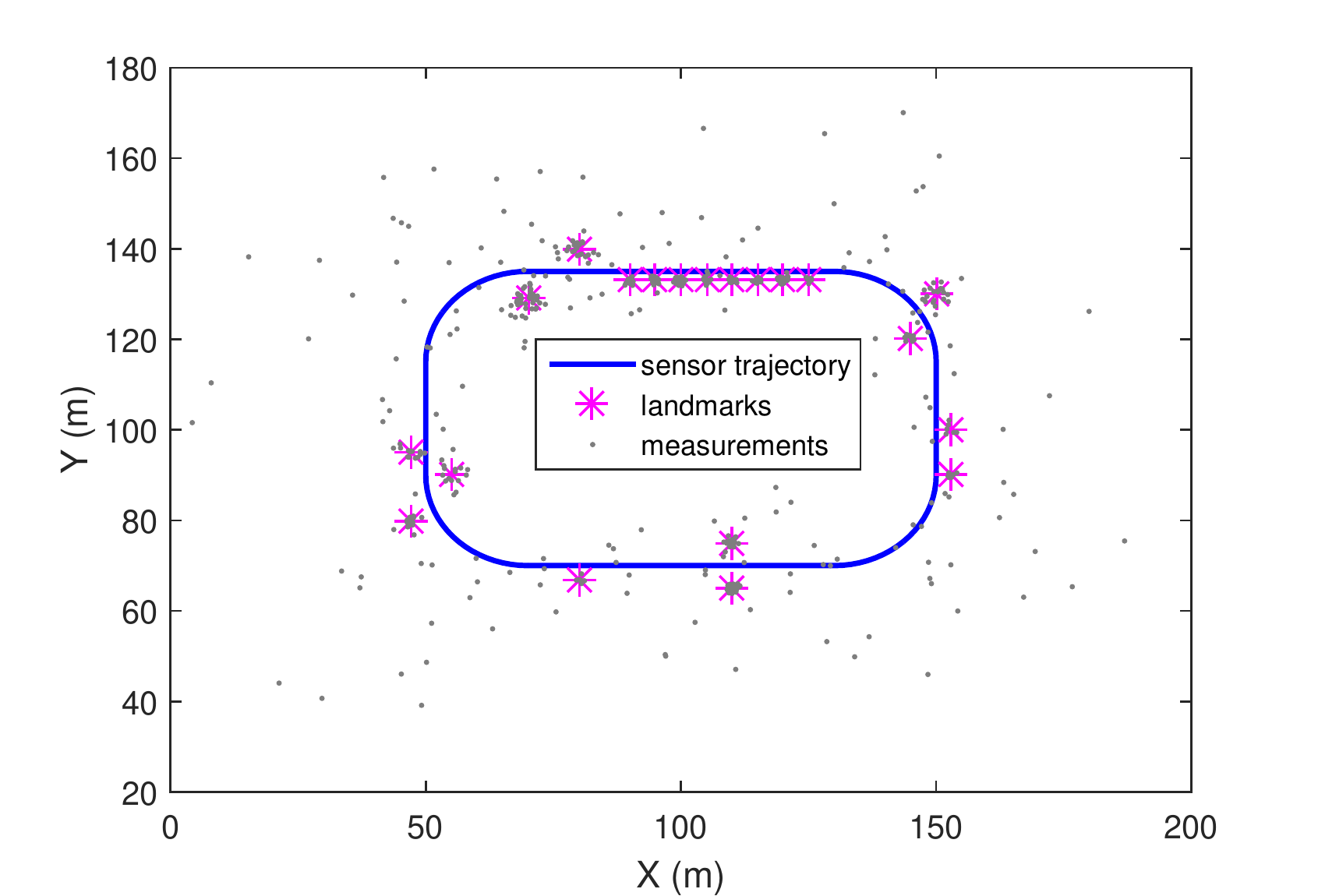}
		 	\caption{The scenario on which the algorithm is evaluated. The vehicle travels one lap over the track and collects 543 measurements in total over 190 time steps.}
		 	\label{fig:scenario}
		 \end{figure}

		 In this section the proposed sampling algorithm is evaluated on a synthetic mapping scenario where a sensor travels over a track and collects measurements. The scenario is depicted in Figure \ref{fig:scenario}. The proposed algorithm is compared to the VBEM mapping method presented in \cite{lundgren2015}.
		  
		  The VBEM method approximates the posterior density of the map by simpler densities and is very computationally efficient; however, this algorithm does not take into account the uncertainty in the number of landmarks. A challenging part of the VBEM is the initialization of the landmarks. The number of landmarks should be set when initializing the algorithm. For mapping a large area one needs to generate a large number of landmarks which cover the AOI reasonably. This is very important as VBEM is sensitive to initialization. The VBEM is guaranteed to find a local optimum which may be far from the global optimum. The Gibbs sampling method samples from the posterior and considers the uncertainties in the number of landmarks at a higher computational price. This algorithm is expected to converge to the true posterior given enough number of MCMC samples. 
		  
		  We use the integrated squared error (ISE) \cite{williams2003} as performance measure, which is defined as the squared L2 norm of the difference of two functions. In our problem, the first function is $\zeta({\mathbf{\theta}})$, the non-normalized Gaussian mixture formed by the true map, and the second is $\zeta(\hat{\mathbf{\theta}})$. The ISE is given by
		 \begin{align}
		 	J({\mathbf{\theta}},\hat{\mathbf{\theta}}) =||\zeta({\mathbf{\theta}})-\zeta(\hat{\mathbf{\theta}}) ||^2_2 = \int (\zeta({\mathbf{\theta}})-\zeta(\hat{\mathbf{\theta}}))^2 d \mathbf{\theta}.
		 	\label{eq:ISE}
		 \end{align}
		 Here, $\mathbf{\theta}$ is formed by an arbitrary order of the true landmarks. This measure of performance accounts for all properties of the map, i.e., positions, extents and weights of landmarks. Furthermore, in \cite{Crouse2011} it is shown that (\ref{eq:ISE}) has a simple and exact form for Gaussian mixtures, which is described as
		 \begin{align}
		 J({\mathbf{\theta}},\hat{\mathbf{\theta}}) &= J_{TT}-2J_{TE}+J_{EE} \\
		 J_{TT} &= \sum_{l_1=1}^{N_T}\sum_{l_2=1}^{N_T} \omega_{l_1}\omega_{l_2}\mathcal{N}(\mathbf{\mu}_{l_1};\mathbf{\mu}_{l_2},\mathbf{\Sigma}_{l_1}+\mathbf{\Sigma}_{l_2}) \\
		 J_{TE} &= \sum_{l_1=1}^{N_T}\sum_{l_2=1}^{N_E} \omega_{l_1}\hat{\omega}_{l_2}\mathcal{N}(\mathbf{\mu}_{l_1};\hat{\mathbf{\mu}}_{l_2},\mathbf{\Sigma}_{l_1}+\hat{\mathbf{\Sigma}}_{l_2}) \\
		 J_{EE} &= \sum_{l_1=1}^{N_E}\sum_{l_2=1}^{N_E} \hat{\omega}_{l_1}\hat{\omega}_{l_2}\mathcal{N}(\hat{\mathbf{\mu}}_{l_1};\hat{\mathbf{\mu}}_{l_2},\hat{\mathbf{\Sigma}}_{l_1}+\hat{\mathbf{\Sigma}}_{l_2}), 
		 \end{align}
		 where $N_T$ is the true number of landmarks.
		 
		 The following parameter values have been used in the simulations. The prior values for different parameters of the map are set to $\mathbf{S}_j^0=5\mathbf{I}$, $\nu_j^0=5$ and $\alpha_0=0.1$, $\beta_0=0.2$. The covariance of the measurement noise is set to $\mathbf{R} = 0.01 ^2\mathbf{I}$ and $\lambda_c=1$. The sensor's field of view has a range of $60$m and an angle of $\pm 30^{\circ}$. The probability of detection is set to one inside the field of view and zero outside.

		 The true map consists of 20 landmarks. These landmarks, together with clutter, have generated 543 measurements over 190 time scans, therefore, there are $10^{931.111}$ ways to partition this set into non-empty cells. The sampling algorithm is run for 120000 iterations, each taking on average 0.2 seconds on a desktop computer with Windows 7 professional, Intel(R) core(TM) i5 CPU 650@ 3.20 GHZ, and 8 GB of RAM. We have chosen a high number of iterations to ensure that we avoid the burn-in period of the sampler. Our results indicate that the algorithm converges earlier (see below). 
		 
		 {We initialize the algorithm with one cell per measurement. This corresponds to each detection being an individual landmark or clutter, depending on the probability of existence. Since this setting corresponds in general to a very unlikely configuration (under the posterior), we considered a burn-in period to remove the contribution of the initial samples of the algorithm. We chose this initialization scheme to show that our algorithm does converge even from a poor starting point.}

\begin{figure}
		  	\centering
		  	\includegraphics[width=0.5\textwidth]{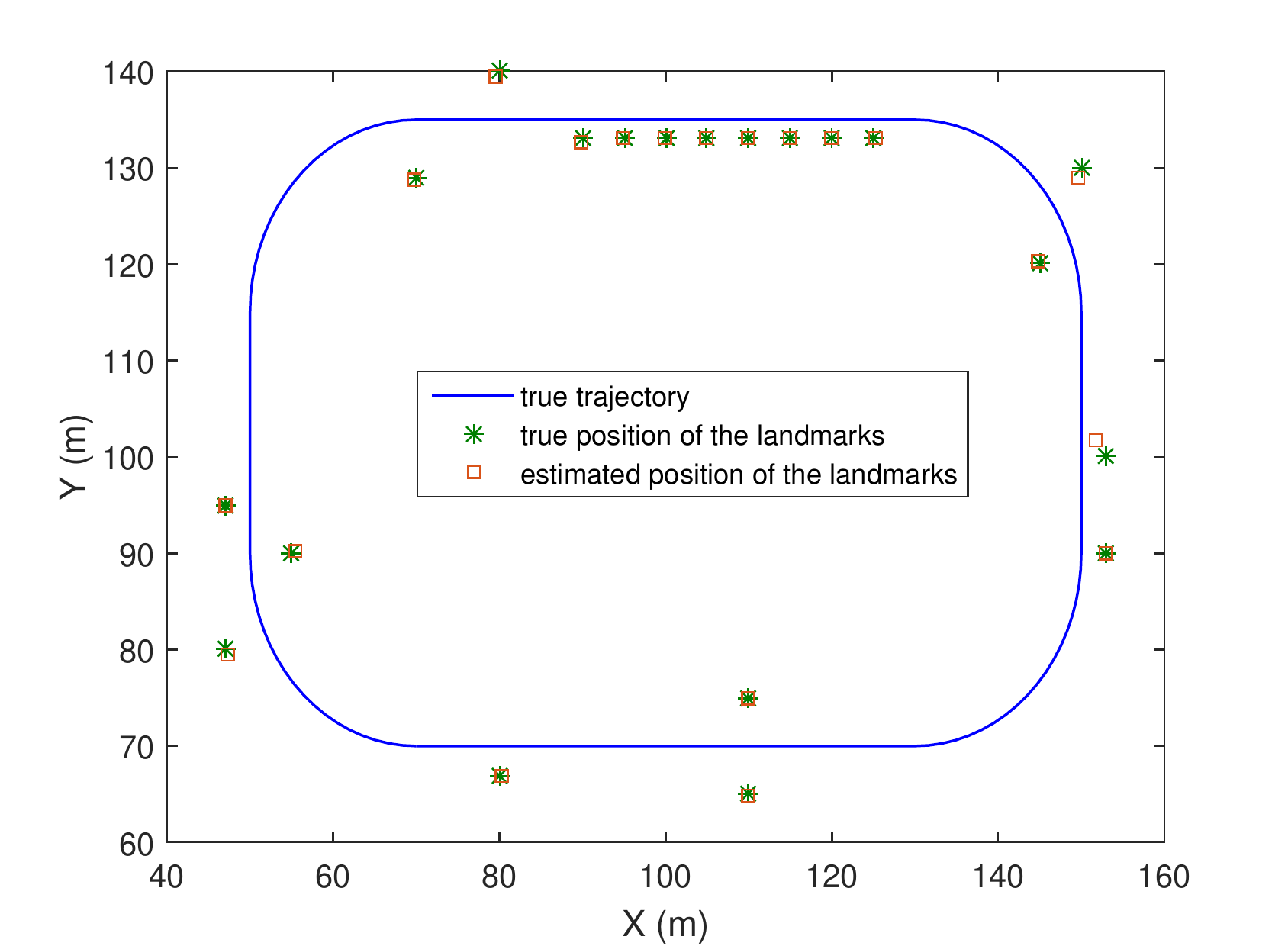}
		  	\caption{Comparison between the true and the estimated positions of the landmarks for the Gibbs sampling method.}
		  	\label{fig:positionComparison}
		  \end{figure}

\begin{figure}
		  	\centering
		  	\includegraphics[width=0.5\textwidth]{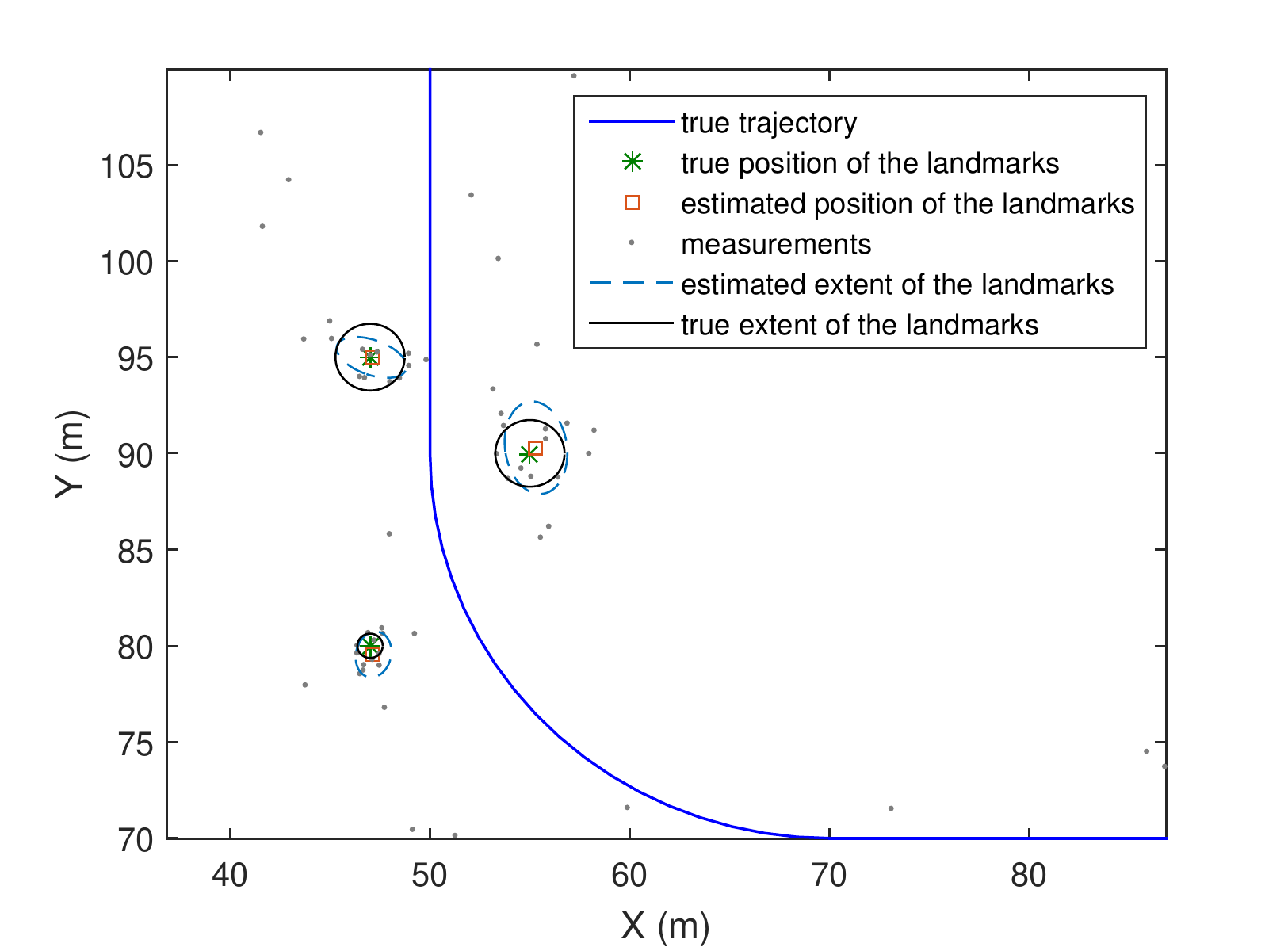}
		  	\caption{A magnified area of the map estimated by the Gibbs sampling method, depicting the measurements, the true and the estimated landmarks. }
		  	\label{fig:zoomedInMap}
		  \end{figure}

\begin{figure}
	\centering
	\includegraphics[width=0.5\textwidth]{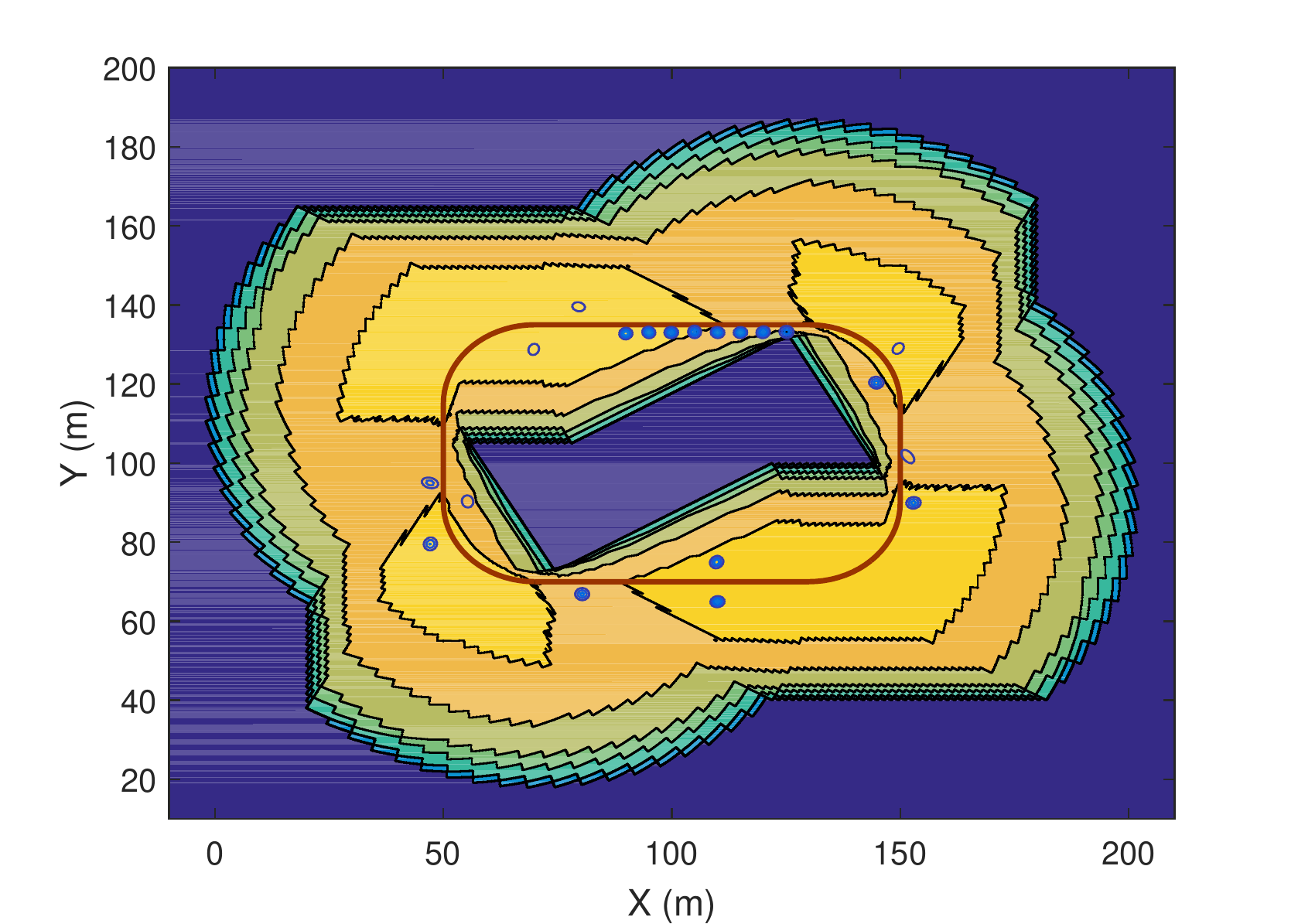}
	\caption{The posterior Poisson rate of the undetected landmarks and the estimated detected landmarks illustrated in one figure. The background is the posterior Poisson rate of the undetected landmarks, where the brighter the color the more frequently the area has been in the FOV and, consequently, the smaller the expected number of undetected landmarks. {Accordingly, the darker the background, the less frequently the area has been observed by the sensor (covered by the FOV over multiple scans) and the higher the Poisson rate.} On top of the background, we have depicted the sensor trajectory and the detected landmarks, marked by ellipses corresponding to their extent.}
	\label{fig:contourPlot}
\end{figure}

The map is estimated by averaging over the final 40000 samples of the chain. In each sample of the map, the cells with existence probability higher than 0.5 are considered as landmarks. The estimated position of the landmarks is compared to the true positions in Figure \ref{fig:positionComparison}. It can be seen that most of the estimated positions are very close to the true ones. In Figure (\ref{fig:zoomedInMap}), an area from the lower left corner of the map has been magnified to illustrate the true map, the measurements and the estimated map of the area. Figure \ref{fig:contourPlot} depicts the posterior Poisson rate of the undetected landmarks calculated according to (\ref{eq:rateU}), the sensor trajectory and the detected landmarks.

The estimated expected number of clutter measurements is calculated by averaging over the final 40000 samples of the MCMC chain. In each sample, the single-member cells with probability of existence lower than 0.5 are assumed to be clutter. In addition, for each sample, the total number of clutter cells divided by the number of time steps is equal to the expected number of clutter measurements. Following this averaging method, the estimated expected number of clutter measurements is 0.7626.
		 
\begin{figure}
		 	\centering
		 	\includegraphics[width=0.5\textwidth]{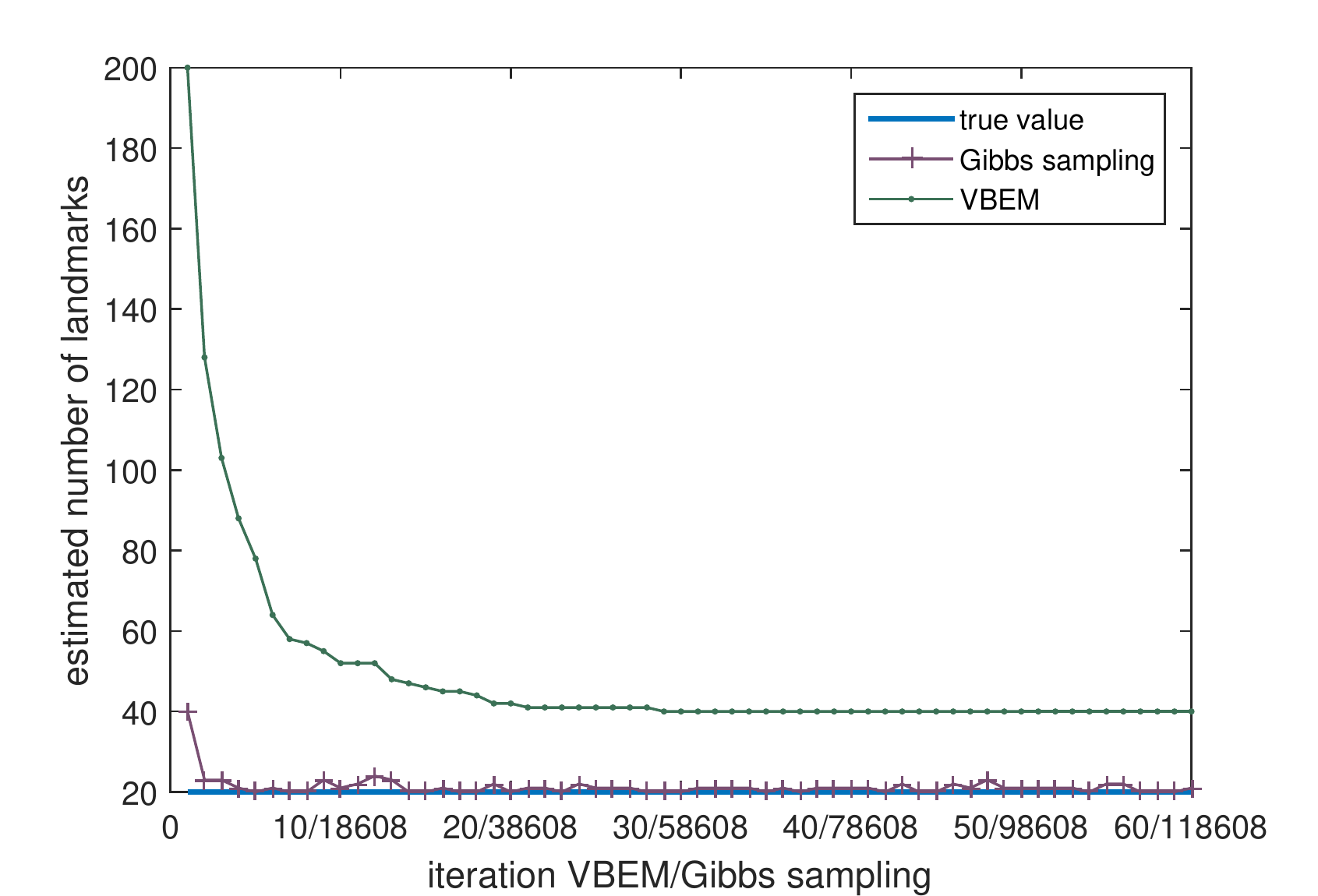}
		 	\caption{The estimated number of landmarks per iteration for VBEM and the proposed algorithm.}
		 	\label{fig:nLandmarkIteration}
		 \end{figure}

Figure \ref{fig:nLandmarkIteration} depicts a comparison of the estimated number of landmarks per iteration between the VBEM method described in \cite{lundgren2015} and our method. The number of landmarks in the VBEM is considered to be those which have weights larger than $0.01$ and in the Gibbs sampling the cells with probability of existence larger than 0.5. It should be noted that the chosen weight threshold for the VBEM is the same as the one used in \cite{lundgren2015} to estimate the number of landmarks. We can see that the Gibbs sampling method provides a more accurate estimate of the number of landmarks. The parameters of the VBEM are set to the same values as \cite{lundgren2015}. The VBEM is initialized with 300 landmarks uniformly distributed over the AOI (recall that the Gibbs sampling method is initialized with 543 single-member cells).   
		  
\begin{figure}
		 	\centering
		 	\includegraphics[width=0.5\textwidth]{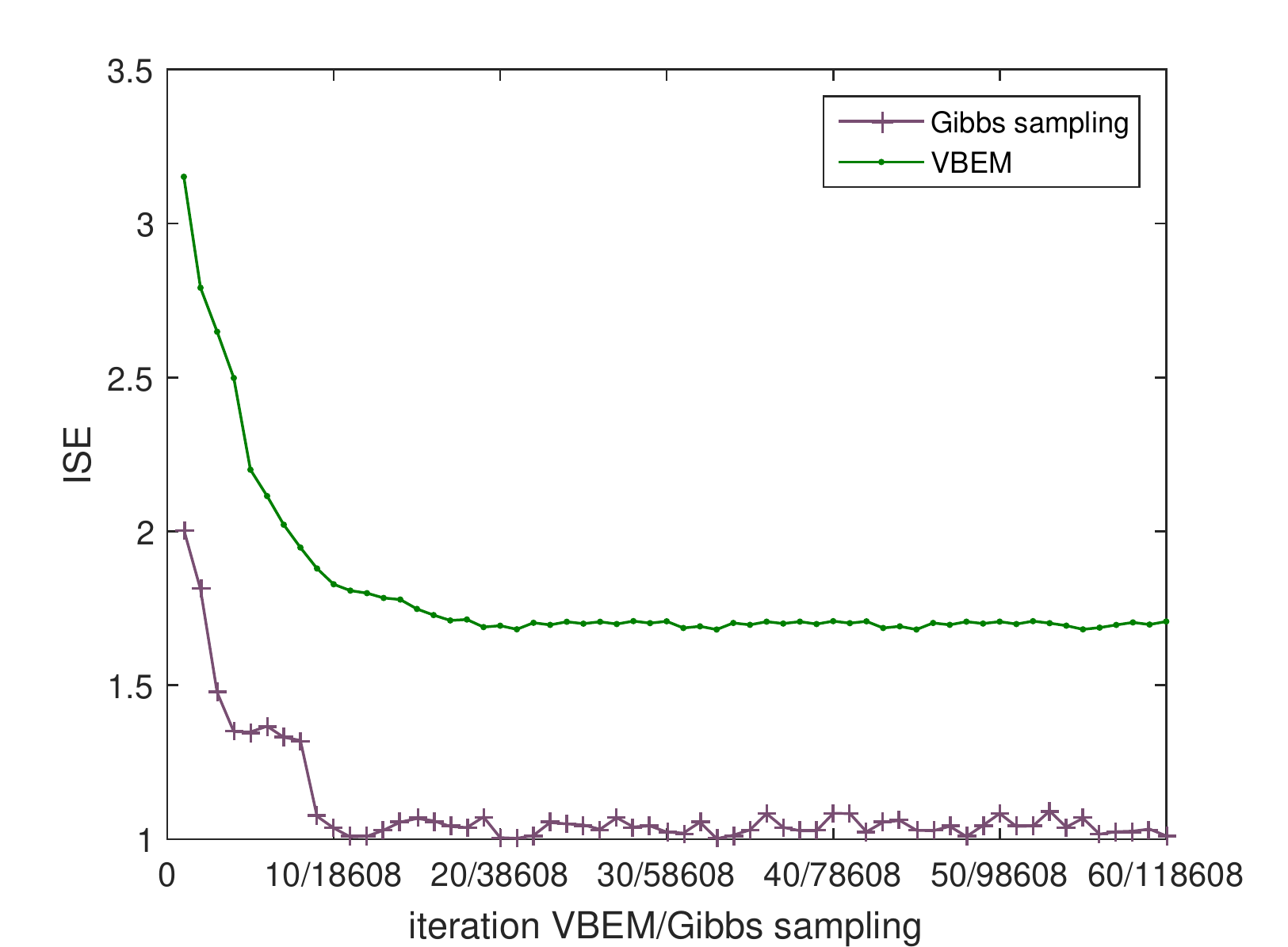}
		 	\caption{Comparison between the ISE of the VBEM and the ISE of the proposed algorithm.}
		 	\label{fig:ISEiteration}
		 \end{figure}

The ISE of the map corresponding to each sample of the MCMC chain is compared to the ISE resulting from each iteration of the VBEM in Figure \ref{fig:ISEiteration}. It can be seen that the Gibbs sampling method has a lower ISE compared to the VBEM. We can see that the Gibbs sampling algorithm provides a more accurate map both in terms of the estimated number of landmarks and the ISE. 
		  
\section{Conclusions}
\label{sec:conclusions}
In this paper, we derive the exact theoretical batch multi-object posterior density of the map using a conjugate prior form described in \cite{granstrom2016gamma}. We propose a solution that uses an extended object model and is capable of handling the uncertainties in the cardinality of the set of landmarks. We use Gibbs sampling to estimate the batch multi-object posterior density of the map. Our method is analogous to well-known sampling approaches in BNPs. The results show that the proposed algorithm can estimate the number of landmarks as well as their properties and outperforms the VBEM method. In addition, we discuss possible parallelization of the proposed algorithm, which makes it suitable for large scale problems. A parallel implementation of the proposed method is a topic for future investigations. {Preliminary results indicate that it is possible to devise a sequential mapping algorithm based on the presented batch mapping algorithm. An interesting future research line is to continue this work and compare a complete and well-designed sequential mapping algorithm with the batch algorithm.
}
\section*{Acknowledgments}
Francisco J.\ R.\ Ruiz acknowledges the EU H2020 program (Marie Sk\l{}odowska-Curie grant agreement 706760) which is supported by ONR N00014-11-1-0651.
\appendices
\section{RFS Background}
\label{sec:RFSbackground}
If the RFS of interest is denoted by $\mathbf{\Theta} = \{\mathbf{\theta}_1,\mathbf{\theta}_2,\ldots,\mathbf{\theta}_n\}$, the RFS density is expressed as
\begin{align}
f(\mathbf{\Theta}) &= p(n) \sum_{\eta}f_n(\mathbf{\theta}_{\eta(1)},\ldots,\mathbf{\theta}_{\eta(n)})
\end{align}  
where $p(n)$ is a cardinality distribution, $f_n(\mathbf{\theta}_{\eta(1)},\ldots,\mathbf{\theta}_{\eta(n)})$ is a cardinality-conditioned joint distribution and $\eta$ accounts for all $n!$ possible permutations ensuring that $f(\mathbf{\Theta})$ is permutation invariant. 

P.g.fls provide an alternative representation of a RFS density, one that is often simpler to deal with. The p.g.fl of a RFS density is a transform of the density defined through the set integral as \cite[p. 371]{mahler2007}
\begin{align}
G[h]&= \int h^{\mathbf{\Theta}} f(\mathbf{\Theta})\delta \mathbf{\Theta}
\end{align}
where $h(\mathbf{\theta})$ is a test function, $h^{\mathbf{\Theta}}\triangleq \prod_{\mathbf{\theta} \in \mathbf{\Theta}}h(\mathbf{\theta})$ and a set integral is
\begin{align}
\int \tau(\mathbf{\Theta})\delta \mathbf{\Theta} &\triangleq \tau(\emptyset) + \sum_{n=1}^{\infty}\frac{1}{n!} \int \tau(\{\mathbf{\theta}_1,\mathbf{\theta}_2,\ldots,\mathbf{\theta}_n\}) d\mathbf{\theta}_1\ldots d\mathbf{\theta}_n.
\end{align}
The RFS density of a union of two independent RFS $\mathbf{\Theta}= \mathbf{\Theta}_1 \bigcup \mathbf{\Theta}_2$ is described by
\begin{align}
f(\mathbf{\Theta})&=\sum_{\mathbf{\Theta}_1 \subseteq \mathbf{\Theta}} f_{\mathbf{\Theta}_1}(\mathbf{\Theta}_1)f_{\mathbf{\Theta}_2}(\mathbf{\Theta}-\mathbf{\Theta}_1)
\end{align}
and the corresponding p.g.fl is \cite[p. 372]{mahler2007}
\begin{align}
G_{\mathbf{\Theta}}[h]&=G_{\mathbf{\Theta}_1}[h]G_{\mathbf{\Theta}_2}[h]. 
\label{eq:pgflUnion}
\end{align}
Additionally, a RFS density can be derived from its p.g.fl by \cite[pp. 375-376, 384]{mahler2007}
\begin{align}
f({\mathbf{\Theta}}) &= \frac{\delta}{\delta {\mathbf{\Theta}}}G[h] |_{h=0} \nonumber \\
&= \frac{\delta^{|\mathbf{\Theta}|}}{\prod_{\mathbf{\theta} \in \mathbf{\Theta}}\delta\mathbf{\theta}}G[h] |_{h=0},
\end{align}
where 
\begin{align}
\frac{\delta}{\delta \mathbf{\theta}}G[h] & \triangleq \lim_{\epsilon \downarrow 0} \frac{G[h+\epsilon \delta_{\mathbf{\theta}}]-G[h]}{\epsilon}, \nonumber
\end{align}
the cardinality of the set $\mathbf{\Theta}$ is denoted by $|\mathbf{\Theta}|$ and $\delta_{\mathbf{\theta}}(\mathbf{\theta}^{\prime})$ is a Dirac delta function centred at $\mathbf{\theta}^{\prime}=\mathbf{\theta}$.

Our model makes use of two random processes, namely, the PPP and the MBM. The RFS density of these processes together with their p.g.fls have been previously derived in \cite{mahler2007} and discussed in \cite{williams2012} \cite{granstrom2016gamma}. 

The RFS density and the p.g.fl of a PPP RFS are described by \cite{mahler2007}
\begin{align}
f(\mathbf{\Theta})&= \exp(-\lambda)\prod_{\mathbf{\theta} \in \mathbf{\Theta}} \lambda f(\mathbf{\theta}) \nonumber \\
G[h] &= \exp(\lambda \langle f ; h \rangle -\lambda)
\end{align}
where $\langle f ; h \rangle = \int f(\mathbf{\theta}) h(\mathbf{\theta}) d\mathbf{\theta}$ and the cardinality of $\mathbf{\Theta}$ is distributed according to a Poisson distribution with rate $\lambda$. Further, the members of $\mathbf{\Theta}$ are independent and identically distributed (iid) according to $f(\mathbf{\theta})$.

A Bernoulli RFS has the following RFS density and p.g.fl,
\begin{align}
f(\mathbf{\Theta})&= \left \{ \begin{array}{cc}
1-r, & \mathbf{\Theta}=\emptyset \\ rf(\mathbf{\theta}), & \mathbf{\Theta}={\mathbf{\theta}} \\ 0, & \text{otherwise}
\end{array} \right. \nonumber \\
G[h]&=1-r+r \langle f;h \rangle 
\label{eq:Bernoulli}
\end{align}
where $r$ is the probability of existence and $f(\mathbf{\theta})$  is the existence-conditioned distribution. 

A multi-Bernoulli process is formed by a union of $N$ independent Bernoulli processes, $\mathbf{\Theta}=\bigcup_{i=1}^N\mathbf{\Theta}_i$. The p.g.fl of this process resulting from (\ref{eq:pgflUnion}) and (\ref{eq:Bernoulli}) is expressed as
\begin{align}
G[h]&= \prod_i (1-r_i+r_i \langle f_i;h \rangle).
\label{eq:pgflMB}
\end{align}   
The RFS density of this process can be described as \cite{williams2012} 
\begin{align}
f(\{\mathbf{\theta}_1,\mathbf{\theta}_2,\ldots,\mathbf{\theta}_n\})&= \sum_{\alpha\in P_N^n} \prod_{i=1}^{N} f_i(\mathbf{\Theta}_{\alpha(i)}),
\label{eq:MBdensity}
\end{align}
where
\begin{align}
P^n_{N} = \{ \alpha: \{1,\ldots,N \} \rightarrow \{0,\ldots,n\}|  \nonumber\\
\{1,\ldots,n\} \subset \alpha(\{1,\ldots,N\}), \nonumber \\
\text{if} \quad \alpha(i)>0, i\neq l \quad \text{then} \quad \alpha(i) \neq \alpha(l) \}
\label{eq:Pmapping}
\end{align}
and
\begin{align}
\mathbf{\Theta}_{\alpha(i)} &= \left \{ \begin{array}{cc}
\emptyset & \alpha(i) = 0 \\ \{\mathbf{\theta}_{\alpha(i)}\} & \alpha(i)>0
\end{array} \right..
\label{eq:Thet_alpha(i)}
\end{align}
That is, $P^n_{N}$ accounts for all different ways of assigning the N component of the multi-Bernoulli to the $n$ landmarks. In addition, each assignment $\alpha$ is a function that maps each Bernoulli component into either a landmark $l \in \{1,2,\ldots,n\}$ or clutter, which is indexed by zero. 

The RFS density of a MBM process is formed by a normalized sum over multi-Bernoulli (MB) RFSs, that is,
\begin{align}
f(\mathbf{\Theta}) = \sum_j W^j f^j(\mathbf{\Theta})
\label{eq:densityMBM}
\end{align} 	
where each weight $W^j$ is related to one data association hypothesis, $\sum_j W^j =1$, and each $f^j(\mathbf{\Theta})$ has the form presented in (\ref{eq:MBdensity}). In addition, the p.g.fl of the MBM process in (\ref{eq:densityMBM}) is formed by a weighted sum of the p.g.fl of MB RFSs described by (\ref{eq:pgflMB}),
\begin{align}
G[h] &= \sum_{j} W^j  \prod _{i} (1-r^{j,i}+r^{j,i} \langle  f^{j,i}; h\rangle).
\end{align}

\subsection{P.g.fl form of multi-object posterior density}
In this section, the p.g.fl of the multi-object posterior is presented for both sequential and batch processing.
The p.g.fl of the sequential posterior density can be calculated using the p.g.fl of the corresponding measurement model. Accordingly, this p.g.fl at time $k$ given the RFS of the measurements up to and including time $k$, $\mathbf{Z}_{1:k}$, is calculated by \cite{mahler2007}
\begin{align}
G_{k|k}[h]&= \frac{\frac{\delta F[g,h]}{\delta \mathbf{Z}_k}|_{g=0}}{\frac{\delta F[\acute{g},\acute{h}]}{\delta \mathbf{Z}_k}|_{\acute{g}=0,\acute{h}=0}}  \label{eq:pgflSeq}\\
F[g,h] & \triangleq \int h^{\mathbf{\Theta}_k}G_k[g|\mathbf{\Theta}_k]f_{k|k-1}(\mathbf{\Theta}_k|\mathbf{Z}_{1:k-1})\delta \mathbf{\Theta}_k \\
G_k[g|\mathbf{\Theta}_k] & \triangleq \int g^{\mathbf{Z}_k} f_k(\mathbf{Z}_k|\mathbf{\Theta}_k)\delta \mathbf{Z}_k
\end{align} 
where $f_{k|k-1}(\mathbf{\Theta}|\mathbf{Z}_{1:k-1})$ is the predicted density of the state. Using this p.g.fl, the posterior density of the RFS of landmarks can be obtained. The complete derivation of this p.g.fl under certain assumptions is given in \cite{granstromArxiv2016} and summarized in Section \ref{sec:sequentialPosterior}.

The batch posterior p.g.fl can be expressed as 
\begin{align}
G_{K|K}[h]&= \frac{\frac{\delta F[g_1,g_2,\ldots,g_K,h]}{\delta \mathbf{Z}_1,\mathbf{Z}_2,\ldots, \mathbf{Z}_K}|_{g_1=0,g_2=0,\ldots,g_K=0}}{\frac{\delta F[\acute{g}_1,\acute{g}_2,\ldots,\acute{g}_K\acute{h}]}{\delta \mathbf{Z}_1,\mathbf{Z}_2,\ldots\delta \mathbf{Z}_K}|_{\acute{g}_1,\acute{g}_2,\ldots,\acute{g}_K,\acute{h}=0}} \label{eq:pgflbatch}
\end{align}
where
\begin{align}
F[g_1,g_2,\ldots,g_K,h]&\triangleq \int h^{\mathbf{\Theta}}G_K[g|\mathbf{\Theta}] f_{0|0}(\mathbf{\Theta})\delta \mathbf{\Theta}
\end{align}
and
\begin{align}
G_K[g|\mathbf{\Theta}]&= \int g_1^{\mathbf{Z}_1} g_2^{\mathbf{Z}_2} \ldots g_K^{\mathbf{Z}_K} \prod_{k=1}^{K}p(\mathbf{Z}_k|\mathbf{\Theta})\delta \mathbf{Z}_1\ldots\mathbf{Z}_K \nonumber \\
&= \prod_{k=1}^{K}G_k[g_k|\mathbf{\Theta}].
\end{align}
Additionally, $p(\mathbf{Z}_k|\mathbf{\Theta})$ is the likelihood function at time $k$ and $f_{0|0}(\mathbf{\Theta})$ is the prior density. These equations are the same as those presented in \cite{mahler2009} in a multi-sensor set-up. Both (\ref{eq:pgflSeq}) and (\ref{eq:pgflbatch}) can be used to derive the batch posterior density.
\section{The batch multi-object posterior density}
\label{sec:appendix}
In this appendix we present the detailed derivation of the batch multi-object posterior for a static map. 

\subsection{The PMBM density}
Since the map consists of two disjoint subsets of detected and undetected landmarks, the RFS density of the map can be written as \cite{mahler2007}
\begin{align}
f({\mathbf{\Theta}}) &= \sum_{\mathbf{\Theta}^d \subset \mathbf{\Theta}} f_d(\mathbf{\Theta}^d)f_u(\mathbf{\Theta} \backslash \mathbf{\Theta}^d).
\end{align}
Additionally, following the results regarding the relation between probability density and p.g.fl in \cite{mahler2007} we can write
\begin{align}
f_d(\mathbf{\Theta}^d) &= \frac{\delta}{\delta \mathbf{\Theta}}G^d[h] |_{h=0}  \nonumber\\
&= \frac{\delta}{\delta \mathbf{\Theta}} (\sum_{j} W_K^j G^{d,j}[h])|_{h=0} \nonumber\\
&= \sum_{j} W_K^j \frac{\delta}{\delta \mathbf{\Theta}}G^{d,j}[h]|_{h=0}
\end{align}
where
\begin{align}
G^{d,j}[h] &= \prod _{i=1}^{I_K^j} (1-r_K^{j,i}+r_K^{j,i} \langle  f_K^{j,i}; h\rangle),
\end{align}
and
\begin{align}
\frac{\delta}{\delta \mathbf{\Theta}}G^{d,j}[h]|_{h=0} & = \sum_{\alpha^j \in P^{|\mathbf{\Theta}^d|}_{I_K^j}} \prod_{i=1}^{I_K^j} f^{j,i}_K(\mathbf{\Theta}_{\alpha^j(i)}).
\end{align}
Similar to (\ref{eq:Pmapping}) and (\ref{eq:Thet_alpha(i)}), the mapping $P^{|\mathbf{\Theta}^d|}_{I_K^j}$ and the components $\mathbf{\Theta}_{\alpha^j(i)}$ are defined as 
\begin{align}
P^{|\mathbf{\Theta}^d|}_{I_K^j} = \{ \alpha^j: \{1,\ldots,I_K^j \} \rightarrow \{0,\ldots,|\mathbf{\Theta}^d|\}|  \nonumber\\
\{1,\ldots,|\mathbf{\Theta}^d|\} \subset \alpha(\{1,\ldots,I_K^j\}), \nonumber \\
\text{if} \quad \alpha^j(i)>0, i\neq l \quad \text{then} \quad \alpha^j(i) \neq \alpha^j(l) \}
\label{eq:mappingMBM}
\end{align}
and 
\begin{align}
\mathbf{\Theta}_{\alpha^j(i)} &= \left \{ \begin{array}{cc}
\emptyset & \alpha^j(i) = 0 \\ \{\mathbf{\theta}_{\alpha^j(i)}\} & \alpha^j(i)>0
\end{array} \right.,
\label{eq:componenrMBM}
\end{align}
where $P^{|\mathbf{\Theta}^d|}_{I_K^j}$ accounts for all different ways of assigning each Bernoulli of each multi-Bernoulli component to a landmark for different number of landmarks. For each multi-Bernoulli, ${|\mathbf{\Theta}^d|}$ cannot exceed $I_K^j$. In addition, $\alpha^j$ is a function that maps each Bernoulli component of the $j$-th multi-Bernoulli into either a landmark $l \in \{1,2,\ldots,{|\mathbf{\Theta}^d|}\}$ or clutter. Similar to (\ref{eq:Bernoulli}), the spatial density of each Bernoulli component is defined as
\begin{align}
f^{j,i}_K(\mathbf{\Theta}) &= \left \{ \begin{array}{cc} 1-r_K^{j,i} & \mathbf{\Theta}=\emptyset \\r_K^{j,i}f_K^{j,i}(\mathbf{\theta}) & \mathbf{\Theta} = \{ \mathbf{\theta}\} \\
0 & \text{otherwise} \end{array} \right..
\end{align}

The Poisson process accounting for the undetected landmarks is given by
\begin{align}
f_u(\mathbf{\Theta} \backslash \mathbf{\Theta}^d) &= e^{-\lambda_K^u} \prod_{\mathbf{\theta}\in \Theta^{u}} \lambda_K^u f_K^u(\mathbf{\theta})
\end{align}
where $\lambda_K^u$ and $f_K^u(\mathbf{\theta})$ are described in (\ref{eq:rateU}) and (\ref{eq:fU}), respectively. 

\subsection{Components of the PMBM density}
The probability of existence of each Bernoulli component in (\ref{eq:rBernouliD}) depends on the cardinality of $\mathbf{C}^{j,i}$. By applying (\ref{eq:PeNew}), (\ref{eq:PeAgain}) and (\ref{eq:PeEmpty}) sequentially we can see that if $|\mathbf{C}^{j,i}| > 1$ then $r_K^{j,i}=1$. Similarly, the probability of existence for $|\mathbf{C}^{j,i}|=1$ in (\ref{eq:rBernouliD}) is derived by by applying (\ref{eq:PeNew}) and (\ref{eq:PeEmpty}), sequentially. 

The spatial density of each Bernoulli component in (\ref{eq:fBernoulliD}) is derived using (\ref{eq:fNew}), (\ref{eq:fAgain}) and (\ref{eq:fEmpty}). 

The Poisson rate and the spatial density of undetected landmarks described in (\ref{eq:rateU}) and (\ref{eq:fU}) are derived by sequentially applying (\ref{eq:rsU}) and (\ref{eq:fsU}), respectively.		
		
		\bibliographystyle{IEEEtran}
		\bibliography{IEEEabrv,refrences}
		
		\vspace{-1.2cm}
		\begin{IEEEbiography}[{\includegraphics[width=1in,height=1.25in,clip,keepaspectratio]{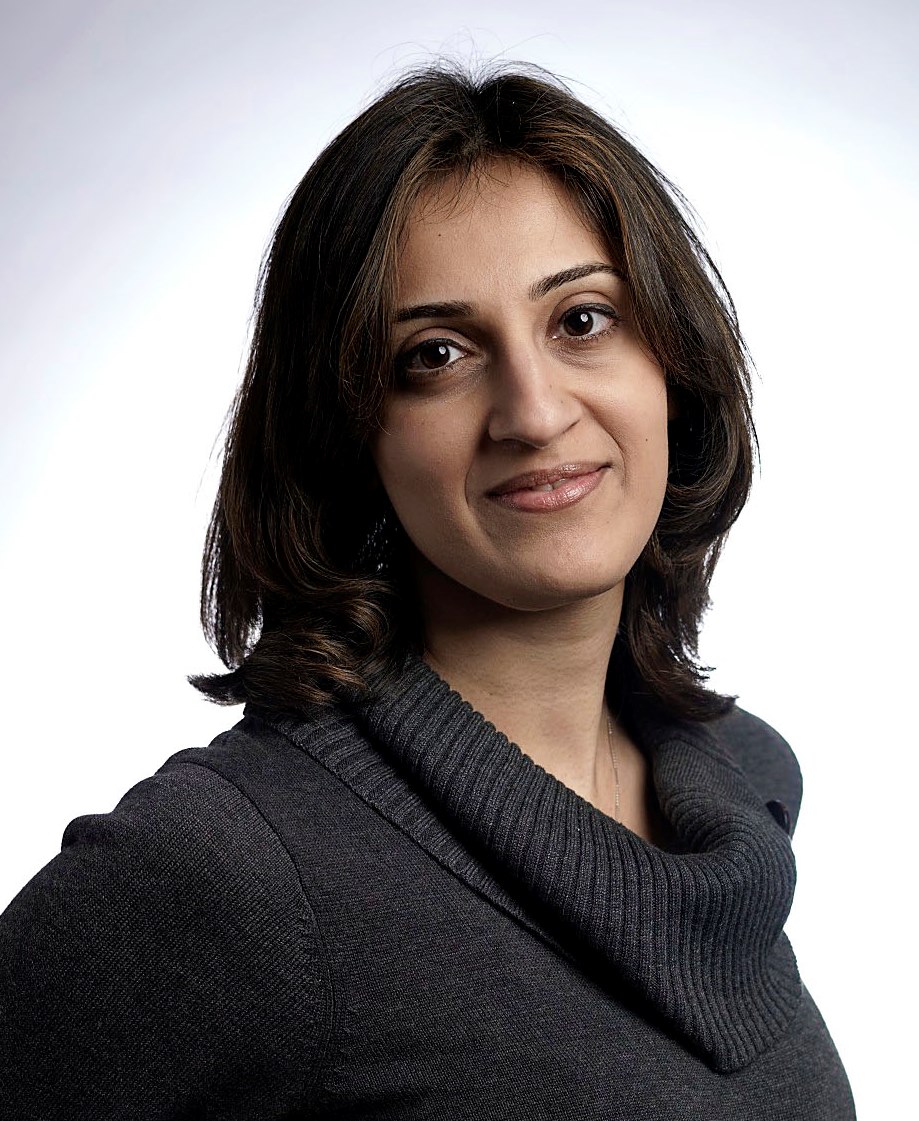}}]{Maryam Fatemi} received the M.Sc. degree in communication systems engineering from AmirKabir University of Technology, Tehran, Iran in 2008, and received the Ph.D. degree in electrical engineering from Chalmers University of Technology, Gothenburg, Sweden in 2016. Since November 2016 she has joined Autoliv Sverige AB.
		Her research interests include Bayesian Inference and nonlinear filtering  with applications to sensor data fusion, autonomous driving and active safety systems.
		\end{IEEEbiography}
		\vspace{-1.5cm}
		\begin{IEEEbiography}[{\includegraphics[width=1in,height=1.25in,clip,keepaspectratio]{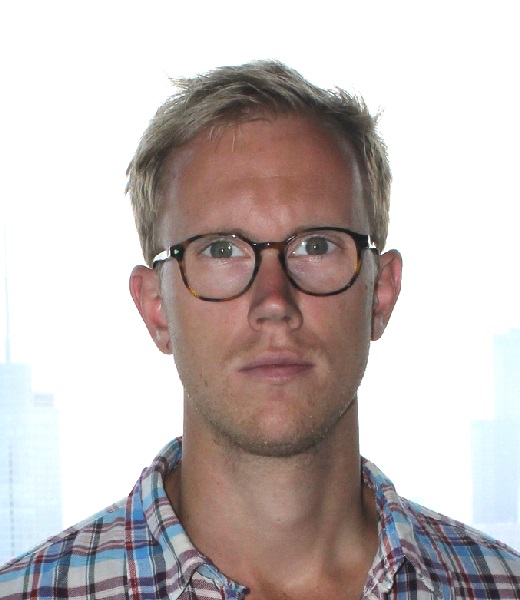}}]{Karl Granstr\"{o}m} (M'08) is a postdoctoral research fellow at the Department of Signals and Systems, Chalmers University of Technology, Gothenburg, Sweden. He received the MSc degree in Applied Physics and Electrical Engineering in May 2008, and the PhD degree in Automatic Control in November 2012, both from Link\"{o}ping University, Sweden. He previously held postdoctoral positions at the Department of Electrical and Computer Engineering at University of Connecticut, USA, from September 2014 to August 2015, and at the Department of Electrical Engineering of Link\"{o}ping University from December 2012 to August 2014. His research interests include estimation theory, multiple model estimation, sensor fusion and target tracking, especially for extended targets. He has received paper awards at the Fusion 2011 and Fusion 2012 conferences.
			\end{IEEEbiography}
			\vspace{-1cm}
			\begin{IEEEbiography}[{\includegraphics[width=1in,height=1.25in,clip,keepaspectratio]{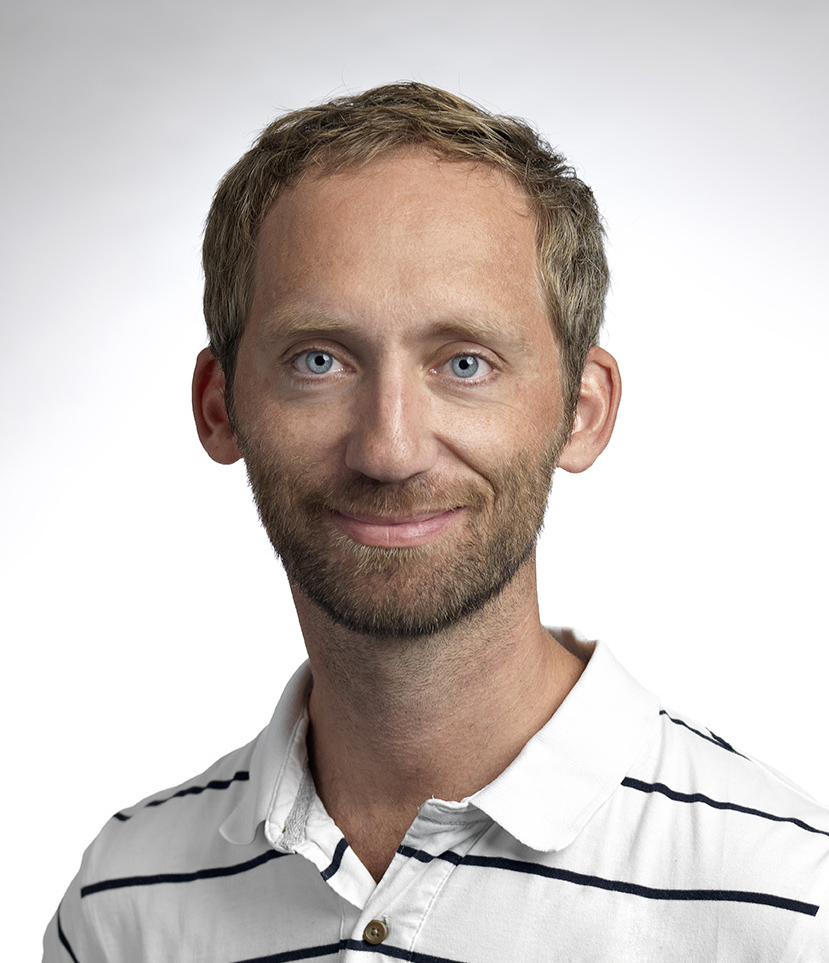}}]{Lennart Svensson} was born in \"{A}lv\"{a}ngen, Sweden in 1976. He received the M.S. degree in electrical engineering in 1999 and the Ph.D. degree in 2004, both from Chalmers University of Technology, Gothenburg, Sweden.
				
				He is currently Professor of Signal Processing, again at Chalmers University of Technology. His main research interests include machine learning and Bayesian inference in general, and nonlinear filtering and tracking in particular.
			\end{IEEEbiography}
			\begin{IEEEbiography}[{\includegraphics[width=1in,height=1.25in,clip,keepaspectratio]{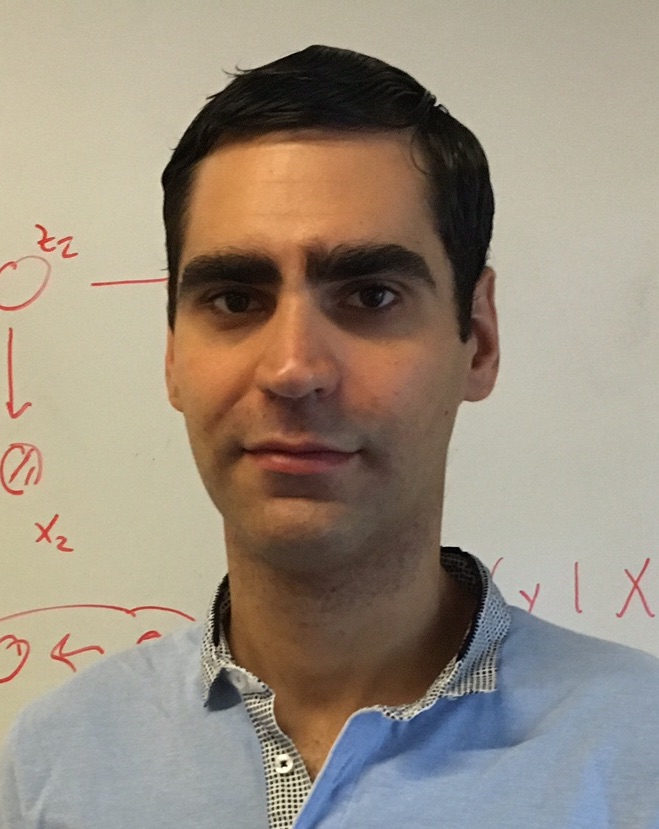}}]{Francisco J. R. Ruiz} is a Postdoctoral Research Scientist who works with David Blei at the Data Science Institute and the Department of Computer Science at Columbia University, and with Zoubin Ghahramani at the Engineering Department of the University of Cambridge. Francisco holds a Marie Sk\l{}odowska-Curie fellowship in the context of the E.U. Horizon 2020 program. He completed his Ph.D. and M.Sc. from the University Carlos III in Madrid. His research is focused on statistical machine learning, specially Bayesian modeling and inference.
			\end{IEEEbiography}
		\vspace{-17cm}
		\begin{IEEEbiography}[{\includegraphics[width=1in,height=1.25in,clip,keepaspectratio]{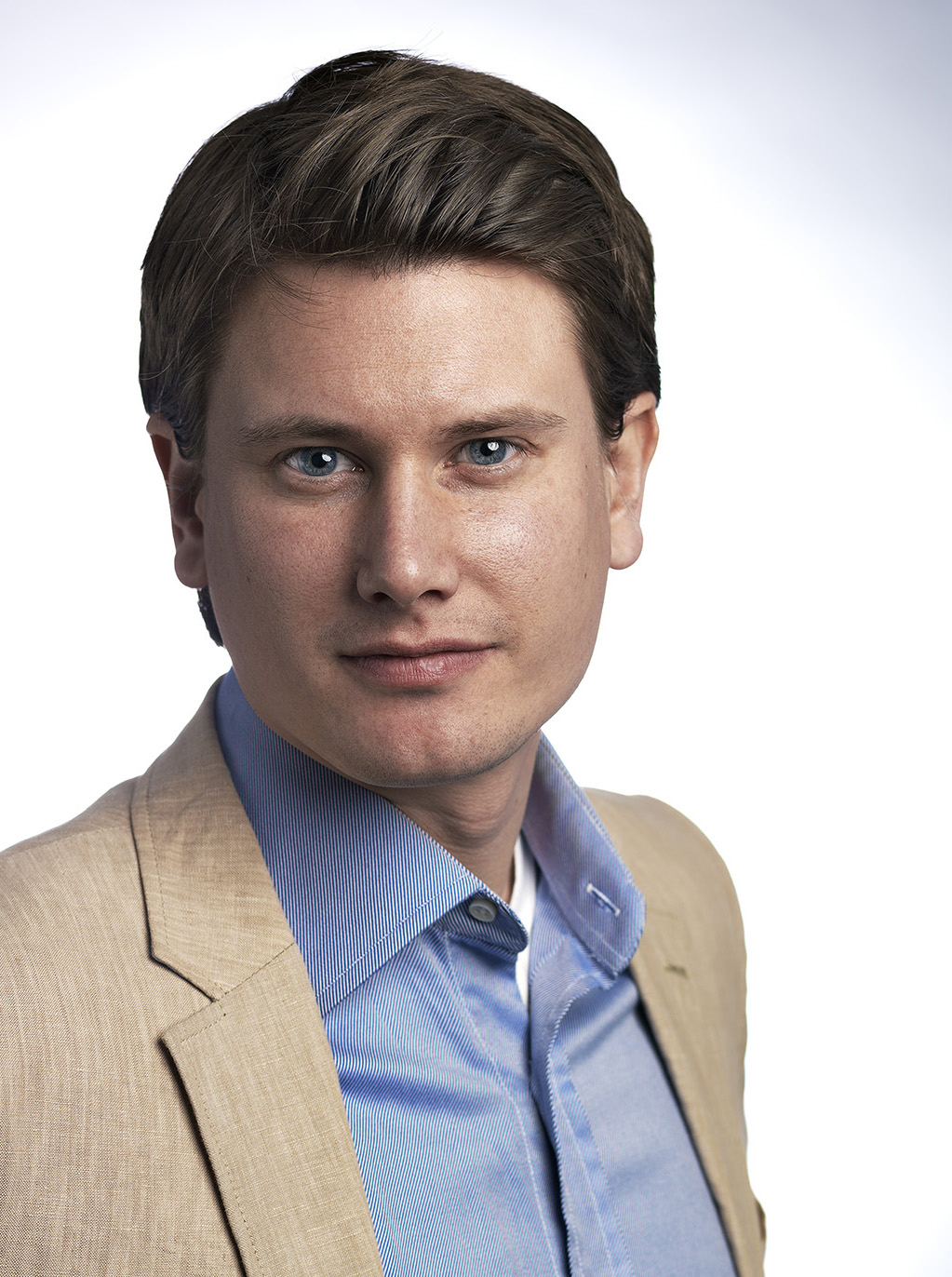}}]{Lars Hammarstrand} was born in Landvetter, Sweden in 1979. He received his M.Sc. and Ph.D. degree in electrical engineering from Chalmers University of Technology, Gothenburg, Sweden, in 2004 and 2010, respectively. 
			
			Between 2004 and 2011, he was with the Active Safety and Chassis Department at Volvo Car Corporation, Gothenburg, conducting research on tracking and sensor fusion methods for active safety systems. Currently, Lars is an Assistant Professor at the Signal Processing group at Chalmers University of Technology where his main research interests are in the fields of estimation, sensor fusion, self-localization and mapping, especially with application to self-driving vehicles.
		\end{IEEEbiography}
		
\end{document}